\documentclass[sn-nature]{sn-jnl}

\usepackage{xr}
\usepackage{xr-hyper}
\usepackage{lmodern}
\usepackage{graphicx}%
\usepackage{multirow}%
\usepackage{amsmath,amssymb,amsfonts}%
\usepackage{amsthm}%
\usepackage{mathrsfs}%
\usepackage[title]{appendix}%
\usepackage{textcomp}%
\usepackage{manyfoot}%
\usepackage{booktabs}%
\usepackage{algorithm}%
\usepackage{algorithmicx}%
\usepackage{algpseudocode}%
\usepackage{listings}%
\usepackage{longtable}
\usepackage{rotating}
\usepackage[utf8]{inputenc} 
\usepackage[T1]{fontenc}    
\usepackage{url}            
\usepackage{booktabs}       
\usepackage{amsfonts}       
\usepackage{graphicx}
\usepackage{natbib}
\usepackage{subcaption}
\usepackage{amsmath}
\usepackage[dvipsnames]{xcolor}
\usepackage{hyperref}

\begin{document}

\title[\#EpiTwitter: Public Health Messaging During the COVID-19 Pandemic]{\#EpiTwitter: Public Health Messaging During the COVID-19 Pandemic}

\author*[1,2]{\fnm{Ashwin} \sur{Rao}}\email{mohanrao@usc.edu}
\equalcont{These authors contributed equally to this work.}

\author[3]{\fnm{Nazanin} \sur{Sabri}}\email{nsabri@ucsd.edu}
\equalcont{These authors contributed equally to this work.}

\author[1,2]{\fnm{Siyi} \sur{Guo}}\email{siyiguo@usc.edu}

\author[4]{\fnm{Louiqa} \sur{Raschid}}\email{lraschid@umd.edu}

\author[2]{\fnm{Kristina} \sur{Lerman}}\email{lerman@isi.edu}

\affil*[1]{\orgdiv{Thomas Lord Department of Computer Science}, \orgname{University of Southern California}, \orgaddress{\city{Los Angeles}, \postcode{90007}, \state{CA}, \country{USA}}}

\affil[2]{\orgdiv{Information Sciences Institute}, \orgname{University of Southern California}, \orgaddress{\city{Marina del Rey}, \postcode{90292}, \state{CA}, \country{USA}}}

\affil[3]{\orgdiv{Department of Computer Science}, \orgname{University of California - San Diego}, \orgaddress{\city{La Jolla}, \postcode{92093}, \state{CA}, \country{USA}}}

\affil[4]{\orgdiv{Institute for Advanced Computer Studies}, \orgname{University of Maryland}, \orgaddress{\city{College Park}, \postcode{20740}, \state{MD}, \country{USA}}}

\maketitle
\begin{abstract}

\textbf{Background:}
Effective communication is crucial during health crises, and social media has become a prominent platform for public health experts to inform and to engage with the public. At the same time, social media also platforms pseudo-experts who may promote contrarian views. Despite the significance of social media, key elements of communication such as the use of moral or emotional language and messaging strategy, particularly during the COVID-19 pandemic, has not been explored.

\noindent \textbf{Objective:}
This study aims to analyze how notable public health experts (PHEs) and pseudo-experts communicated with the public during the COVID-19 pandemic. 
Our focus is the emotional and moral language they used in their messages across a range of pandemic issues. We also study their engagement with political elites and how the public engaged with PHEs to better understand the impact of these health experts on the public discourse.

\noindent \textbf{Methods:}
We gathered a dataset of original tweets from 489 PHEs and 356 pseudo-experts on Twitter (now X) from January 2020 to January 2021, as well as replies to the original tweets from the PHEs. 
We identified the key issues that PHEs and pseudo-experts prioritized. 
We also determined the emotional and moral language in both the original tweets and the replies. 
This approach enabled us to characterize key priorities for PHEs and pseudo-experts, as well as differences in messaging strategy between these two groups. 
We also evaluated the influence of PHE language and strategy on the public response.

\noindent \textbf{Results:}
Our analyses revealed that PHEs focus on masking, healthcare, education, and vaccines, whereas pseudo-experts discuss therapeutics and lockdowns more frequently. PHEs typically used positive emotional language across all issues, expressing optimism and joy. Pseudo-experts often utilized negative emotions of pessimism and disgust, while limiting positive emotional language to origins and therapeutics. Along the dimensions of moral language, PHEs and pseudo-experts differ on care versus harm, and authority versus subversion, across different issues. Negative emotional and moral language tends to boost engagement in COVID-19 discussions, across all issues. However, the use of positive language by PHEs increases the use of positive language in the public responses. PHEs act as liberal partisans: they express more positive affect  in their posts directed at liberals and more negative affect directed at conservative elites. In contrast, pseudo-experts act as conservative partisans. These results provide nuanced insights into the elements that have polarized the COVID-19 discourse.

\noindent \textbf{Conclusion:}
Understanding the nature of the public response to PHE's messages on social media is essential for refining communication strategies during health crises. 
Our findings emphasize the need for experts to consider the strategic use of moral and emotional language in their messages in order to reduce polarization and enhance public trust.

\end{abstract}


\section*{Introduction}
The COVID-19 pandemic created a world-wide public health crisis, disrupting daily lives and overwhelming healthcare facilities. During this time the need for communicating reliable medical information and public health guidance became very important. Social media platforms such as Twitter (now X) provided a space for public health experts (PHEs) in government, academia, and think tanks to communicate timely, reliable information about the pandemic to the public, to discuss research, and to provide guidance~\cite{tang2021important,abbas2021role}. 

However, as the pandemic progressed, discussions around the pandemic grew highly contentious and ideologically polarized \cite{jiang2020political,rao2023pandemic}. 
As public's trust in institutions and experts eroded, health-related misinformation proliferated about all aspects of the pandemic, from its origins to alternative treatments and the efficacy of nonpharmaceutical interventions, and eventually the vaccine~\cite{rao2022partisan}. 
At the heart of this proliferation were influential \textit{pseudo-experts}, like the ``Disinformation Dozen''~\cite{disinfo2021}, who amplified contrarian perspectives and challenged the recommendations made by PHEs.
When PHEs advocated masking and social distancing, pseudo-experts promoted alternative treatments like Hydroxychloroquine and Ivermectin. When PHEs advocated vaccination as a method for reaching herd immunity, pseudo-experts emphasized vaccine injury.

In this paper, we examine the messages shared by influential accounts posting about public health through the lens of affect. In psychology, ``affect'' is the experience of feeling or emotion, and it significantly  shapes individual's attitudes, beliefs and behaviors. In online interactions, affect  influences how a message is crafted and how it resonates with audiences, ultimately affecting the message's spread and impact. Research shows that people respond to the emotions expressed in online messages~\cite{kramer2014experimental}, although due to an asymmetry in human cognition~\cite{baumeister2001bad}, posts expressing negative emotions receive more engagement than positive posts~\cite{coviello2014detecting,ferrara2015measuring}. It has also been shown that emotionally charged messages, especially ones tapping into moral sentiments like outrage, spread farther online~\cite{brady2017emotion,brady2021social}. 
Affect provides reliable indicators for gauging  public response to major events and policy decisions \cite{renstrom2021emotions,sadler2005emotions,grover2019moral,sterling2018moral,sagi2014moral} and interacts with ideology to fuel polarization. In fact, political scientists have identified affective polarization---a phenomenon in which each party likes and trusts members of its own party by dislikes and distrusts members of the other party---as a fundamental threat to effective governance~\cite{iyengar2015fear,iyengar2019origins}.
Public's reactions to the pandemic, as measured via attitudes and sentiments expressed in online messages, were multifaceted~\cite{tsao2021social} and grew polarized early in the pandemic~\cite{jiang2020political}. Moreover, there was an ideological asymmetry wherein conservatives shared more low quality health information than liberals~\cite{rao2021political} and were also exposed to more misinformation~\cite{rao2022partisan}. In addition, conservatives expressed more negative moral sentiments in online posts about the pandemic than liberals~\cite{rao2023pandemic}.
However, to the best of our knowledge, few works have focused on  online influencers and experts who shaped public health policy and disseminated health-related information to the public. As a result, we know little about the messaging strategies they used, the role that affect played in these messages,  
and how the public responded to the messages.

To address these knowledge gaps, we examined messages posted by public health experts and pseudo-experts on Twitter during the COVID-19 pandemic. We identified a set of 489 PHEs and 356 pseudo-experts and collected over 372K original tweets that they posted between January 21, 2020 to January 20, 2021. Collectively, these accounts had a vast reach: each PHE had on average 94K followers (estimated reach around 45M) and pseudo-experts had on average 78K followers (estimated reach around 30M). In addition, we also collected replies to over 19.5K original tweets posted by PHEs during this time period. Our objectives were two-fold: 
Identify \textit{what} public health influencers talk about online, 
and \textit{how} they talk;
(ii) Identify factors that impacted public engagement with the PHEs.  

%

We leverage methods introduced in \cite{rao2023pandemic} to identify  tweets about seven important pandemic-related  issues: origins of the virus, lockdowna and stay-at-home orders, masking mandates, online schooling and education, healthcare, alternative treatments and therapeutics, and vaccines. We use state-of-the-art classifiers~\cite{alhuzali2021spanemo,guo2023data} to identify emotional and moral language used  tweets. We then use regression to compare how  affect shapes the health-related messages on different issues posted by PHEs and pseudo experts. Finally, we collect all replies for a sample of PHE tweets to study how the use of emotional and moral language impacts engagement by the public with these messages. These analyses yield the following key findings:
\begin{enumerate}
    \item PHEs focus on masking, healthcare, education, and vaccines, whereas pseudo-experts discuss therapeutics and lockdowns more frequently.
    \item PHEs typically used positive emotional language across all issues, expressing optimism and joy. 
    Pseudo-experts often utilized negative emotions of pessimism and disgust, while limiting positive emotional language to origins and therapeutics.
    \item Along the dimensions of moral language, PHEs and pseudo-experts differ on care versus harm, and authority versus subversion, across different issues.
     \item Negative emotional and moral language tends to boost engagement in COVID-19 discussions, across all issues. However, the use of positive language by PHEs increases the use of positive language in the public responses.
    \item PHEs act as liberal partisans: they express more positive affect  in their posts directed at liberals and more negative affect directed at conservative elites. In contrast, pseudo-experts act as conservative partisans. 
\end{enumerate}

Our study examines the inherent complexities in public health communication dynamics during the pandemic. Our analysis reveals disparities in not only the issues promoted by public health entities and pseudo experts who advance contrarian perspectives but also, highlight differences in the use of emotional and moral language along these issues. Furthermore, our observations reveal asymmetries in emotional and moral language directed towards political elites, indicative of underlying ideological schisms amongst the scientific elite. This diverges from established practices that establish the efficacy of positive framing techniques for fostering consensus. These disparities and asymmetries may contribute to exacerbate public distrust towards public health messaging.




\section*{Related Work}

\subsection*{Public Health Messaging during COVID-19}

Previous literature \cite{slothuus2021political,bullock2020party} shows that the public follows cues from in-group elites and often oppose cues from the out-group. Effective messaging strategies can be crucial in times of public health crises. Individuals with higher COVID-19 knowledge practiced more protective behaviors \cite{milich2024effective,kim2021impact}. Messaging that highlights risks to younger adults, in addition to risks to older adults was found to bring about a higher threat perception about COVID-19 \cite{utych2020age}. On the other hand, messaging that appeals to the audience's morals or fears for compliance was found to be polarising, divisive and undermined social cohesion \cite{mcclaughlin2023reception}. Latino vaccination rates saw significant increases in one Maryland County following the adoption of a cartoon grandmother \cite{wapo2021vaxx}. Messaging comprised of vaccine safety and/or efficacy information and political leaders' endorsement of vaccination were found to be highly effective strategies \cite{xia2023motivating}. Unvaccinated \cite{pink2021elite} found that Republicans who were exposed to enodrsements from Republican elites witnessed higher vaccination intentions than those who viewed the Democratic elite endorsement with out-group elite exposures proving counter-productive.


\subsection*{COVID-19 Pseudo-Science and Conspiracy Theories}

During the COVID-19 pandemic, social media was used by a large number of PHEs and healthcare providers to communicate best practices and information to the people, as well as to gain quicker access to healthcare information \cite{goel2020social}. The pandemic, however, was a multifaceted public health crisis, presenting society with unprecedented challenges for which there was no established playbook. This coupled with the polarization \cite{rao2021political, rao2022partisan, jiang2021social} of the pandemic laid bare a fractured public health messaging apparatus \cite{box2021meaningful,green2020elusive,van2021public,noar2020mis}. The emergence of contradictory theories and two polarized groups of influential elites and experts \cite{mabrey2021disinformation,nogara2022disinformation}, conspiracy theories about the origins of the pandemic, its severity and, the efficacy of prophylactic measures started to take hold \cite{douglas2021covid,bavel2020using}. Initial theories revolved around the severity of the virus with several calling it a ``hoax'' and ``plandemic'' \cite{nyt2020plandemic}. A study by the Pew Research Center \cite{pew2020conspiracy} found that nearly 25\% of the survey responders believed in the conspiracy theory propagated online that COVID-19 was probably created intentionally by powerful people. Another study found that nearly 3 in 10 Americans believe that COVID-19 was artificially created in a lab \cite{pew2020lab}. Theories about virus transmission being connected to 5G, bats, pangolins and wet-markets were widely propagated by conspiracy theorists on social media \cite{langguth2023covid,taylor2020bats}. As the pandemic progressed, we also witnessed the propagation of pseudo-scientific cures for COVID-19 \cite{huff2020hcq,nyt2021ivermectin}. With increased pandemic related engagement from the general public, these conspiracy theories soon started to proliferate on social media platforms \cite{desai2022misinformation, kearney2020twitter,ahmed2020covid, gruzd2020going,ferrara2020types,gottlieb2020information}. \cite{antonakis2021leadership} highlights the role of influential accounts in mitigation efforts. As influential elites, often holding advanced medical degrees began contradicting with other PHEs on various aspects of the pandemic, consensus amongst the general public suffered as a consequence and at times resulted in grave outcomes \cite{geleris2020observational,sehgal2022association,pradelle2024deaths}.

\subsection*{Emotions and Moral Language Use During COVID-19}

Expression of fear and anger were found to indicate support for restrictive COVID-19 mitigation policies such as lockdowns to limit the spread of COVID-19 \cite{renstrom2021emotions}, while anxiety predicted support for economic policies. Anger was found to indicate support for aggressive responses to transgressors \cite{sadler2005emotions,skitka2006confrontational}. \cite{hatemi2013fear} found fear to be a strong underlying factor in anti-immigration and pro-segregation stances. Previous studies \cite{apa2021stress} relied on surveys to show an increase in distress and uncertainty during the pandemic. \cite{agrawal2022covid} investigated sentiments of Tweets about the COVID-19 vaccine, post-Covid health factors, and health service providers. Among the three topics, healthcare providers had the largest positive sentiment, resulting in an inference that posters were happy with their care and appreciated the work of health providers. \cite{lwin2020global} found that public emotions in Twitter shifted from fear to anger early in the pandemic. \cite{wheaton2021fear} revealed that greater susceptibility to emotion contagion was associated with greater concern about the spread of COVID-19. The care and fairness moral foundations were found to correlate with compliance of COVID-19 health recommendations including masking, staying at home and social distancing \cite{diaz2022reactance,chan2021moral}. Moral attitudes were also able to predict county-level vaccination rate \cite{reimer2022moral} and vaccine hesitancy \cite{nan2022moral}. \cite{pacheco2022holistic} found that care/harm was associated with pro-vaccine sentiment, whereas liberty/oppression was correlated with anti-vaccine attitudes. While vaccinations are a critical polarizing issue in the discussion of COVID-19, no work as of yet has explored differences in moral appeals across a broader range of contentious COVID-19 issues.

\subsection*{COVID-19 Engagement on Social Media}


Social sharing of opinions and emotions is ubiquitous, and social media has greatly expanded its scope~\cite{john2013social,bazarova2015social}. \cite{bazarova2015social} investigated how responses to what a user shared affected their feeling of satisfaction. The authors argue that while sharing emotions with large groups of users online is often viewed as over-sharing, it is likely a means to increase the likelihood of receiving responses from virtual listeners. Analyzing Facebook status updates, \cite{burke2016once} found that posts with positive emotions received more likes. The comments associated with these posts were also more positive \cite{burke2016once}. Positive emotion words were also shown to have a positive correlation with the number of retweets \cite{kim2012role}. \cite{sousa2010characterization} reports that while social connections dominate reply behavior, for authors with large ego networks, there is a separation between who replies based on the topic of the post. Early thematic analysis of public replies to the pandemic found themes of prevention, symptoms, views on politicians and, humor \cite{bojja2020early}. Replies by anti-vaxx users were found to be more toxic than users with other beliefs about vaccines \cite{miyazaki2021characterizing}. \cite{gallagher2021sustained} found groups to preferentially amplify elites that are demographically similar to them. 

\section*{Data and Methods}

We begin by describing our data collection procedure, and present statistics and basic characteristics of the dataset. We believe that this description provides the reader with additional insights to better interpret the results. Finally, we describe our content analysis procedure and models used to produce results. 

\subsection*{Identifying Public Health and Pseudo Experts}
We identified a seed set of 30 PHEs and 30 pseudo-experts who were active on social media platforms during the pandemic (see Table~\ref{tab:expert_seed_list}. While the seed set of PHEs comprised largely of individuals who hold advanced degrees in medicine, epidemiology, genomics, infectious diseases or vaccine development, allowing them to offer well-informed, scientifically grounded perspectives on the pandemic, pseudo-experts include individuals with or without medical degrees, proposing pseudo-scientific theories, alternative treatments and therapeutics. The group also includes the ``Disinformation Dozen'', a group of individuals and organizations identified by the Center for Countering Digital Hate as being responsible for promoting false claims about COVID-19~\cite{disinfo2021}.     

\begin{table}[!ht]
\centering
\begin{tabular}{p{1.in}p{3.2in}}
\toprule
Groups &Twitter Handles \\
\midrule
Public Health Experts (PHEs) & EricTopol, PeterHotez, ashishkjha, trvrb, EpiEllie, JuliaRaifman, devisridhar, meganranney, luckytran, asosin, DrLeanaWen, dremilyportermd, DrJaimeFriedman, davidwdowdy, BhramarBioStat, geochurch, DrEricDing, michaelmina\_lab, Bob\_Wachter, JenniferNuzzo, mtosterholm, MonicaGandhi9, cmyeaton, nataliexdean, angie\_rasmussen, ProfEmilyOster, mlipsitch, drlucymcbride, ScottGottliebMD, CDCDirector, Surgeon\_General \\
\hline
Pseudo Experts &  mercola, LEEHIEB\_MD, stella\_immanuel, DrOz, DrThomasLevy, DrJudyAMikovits, va\_shiva, Drericnepute1, DrButtar, DrArtinMassihi, davidicke, mrmarksteel, drscottjensen, cameronks, RobertKennedyJr, TyCharleneB, BusyDrT, IslamRizza, unhealthytruth, sayerjigmi, kellybroganmd, DrChrisNorthrup, DrBenTapper1, DrZachBush, SherrillSellman, AFLDSorg, DrSimoneGold, jennybethm, drcole12, JamesTodaroMD, Covid19Critical, DrJohnWitcher \\
\bottomrule
\end{tabular}
\caption{Twitter Handles. Twitter Handles of Accounts Associated with Public Health Experts and Pseudo Experts}
\label{tab:expert_seed_list}
\end{table}

\begin{figure}[!htb]
    \centering
   \includegraphics[width=0.6\textwidth]{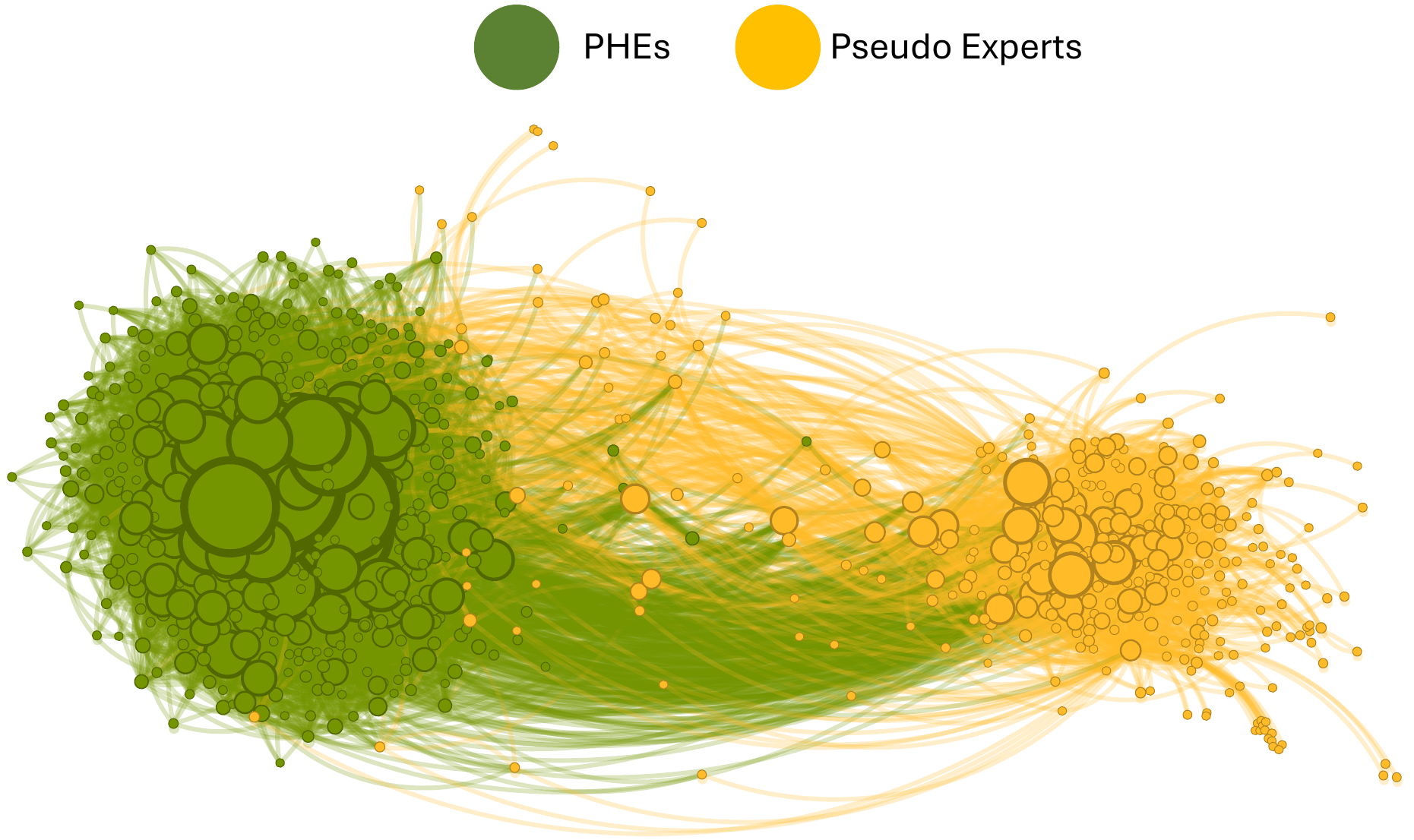}
    \caption{Retweet interactions. Nodes represent PHEs (green) and pseudo-experts (orange) and retweet interactions between them. Green edges represent interactions where a PHE was retweeted, and orange edges represent interactions where a pseudo-expert was retweeted. The size of the node is proportional to the number of times the expert was retweeted.}
    \label{fig:rt_network}
\end{figure}

In order to expand these sets of experts, we rely on a publicly available COVID-19 Twitter dataset \cite{chen2020tracking} comprising of over 1B COVID-19 tweets gathered between January 21, 2020 to January 20, 2021. Previous studies have shown that retweet interactions where users repost or reshare content originally generated by other users, indicates an endorsement of the content \cite{metaxas2015retweets,boyd2010tweet}. Working with the assumption that individuals often agree with others who share identical beliefs \cite{rao2022partisan,cinelli2021echo,colleoni2014echo,kang2012using}, we isolate retweet interactions involving PHEs and pseudo-experts to construct two networks --- one involving accounts retweeted by PHEs and the other  accounts retweeted by pseudo-experts.  We use Eigenvector centrality~\cite{ghosh2010predicting} to rank influential accounts who are retweeted by at least one PHE or pseudo-expert, and select the top-500 most influential accounts. We remove organizations from this list, which leaves us with $489$ and $356$ accounts corresponding to individual PHEs and pseudo-experts respectively. Having identified the group of health and pseudo-experts, we then extract their tweets. Overall, we have $340K$ tweets from PHEs and $175K$ tweets from pseudo-experts. 

Figure \ref{fig:rt_network} shows the retweet interactions network between PHEs and pseudo-experts ($845$ nodes and $107K$ edges). The color of the edge is dependent on the target node. Green edges represent interactions where a PHE was retweeted, whereas orange edges represent interactions where a pseudo-expert was retweeted. The size of the node is proportional to how many times the account was retweeted: highly retweeted experts have larger node sizes. The network shows two tightly knit communities, one for each group, with sparse between-community interactions. This structure is typical of online echo chambers and suggests that each community mainly listens to their own community.



\begin{figure}[!htb]
    \centering
    \begin{subfigure}[!h]{0.49\textwidth}
    \includegraphics[width=\textwidth]{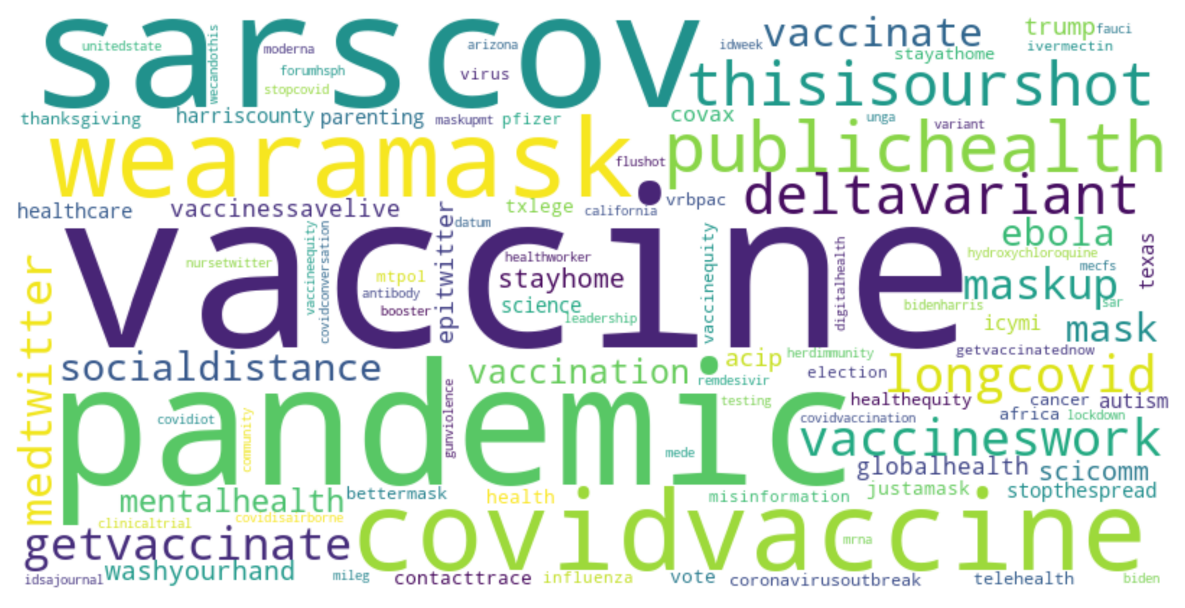}
    \caption{PHEs} 
    \end{subfigure}
    \begin{subfigure}[!h]{0.49\textwidth}
    \includegraphics[width=\textwidth]{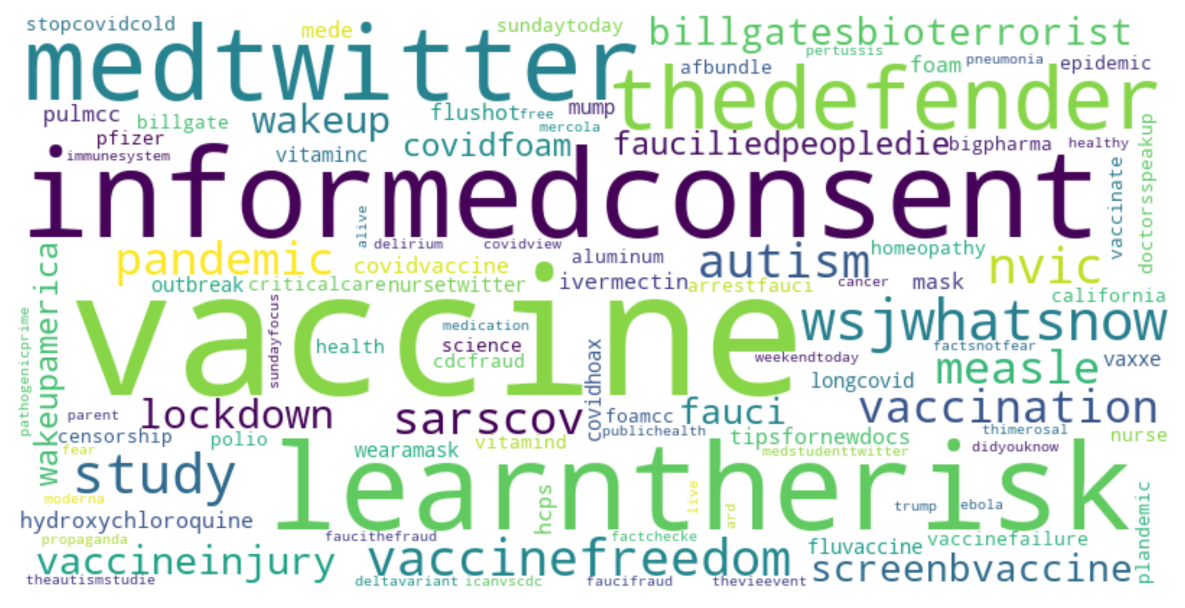}
    \caption{Pseudo Experts} 
    \end{subfigure}
    \caption{Hashtag usage. Wordclouds highlight the most prominent 100 hashtags used by PHEs and pseudo-experts on Twitter.}
    \label{fig:phe_hashtags}
\end{figure}

Figure~\ref{fig:phe_hashtags} shows the wordclouds of 100 most popular hashtags used by PHEs and pseudo-experts. There are notable similarities and differences. While ``vaccine'' is the most important topic for both groups, PHEs unsurprisingly mention ``vaccineswork'' and ``vaccinate,'' in contrast to tweets from pseudo-experts that mention ``vaccineinjury'', ``vaccinefreedom'' and urge people to ``learntherisk'' of vaccines.

\begin{figure}[!htb]
    \centering
    \begin{subfigure}{0.49\textwidth}
    \includegraphics[width=\textwidth]{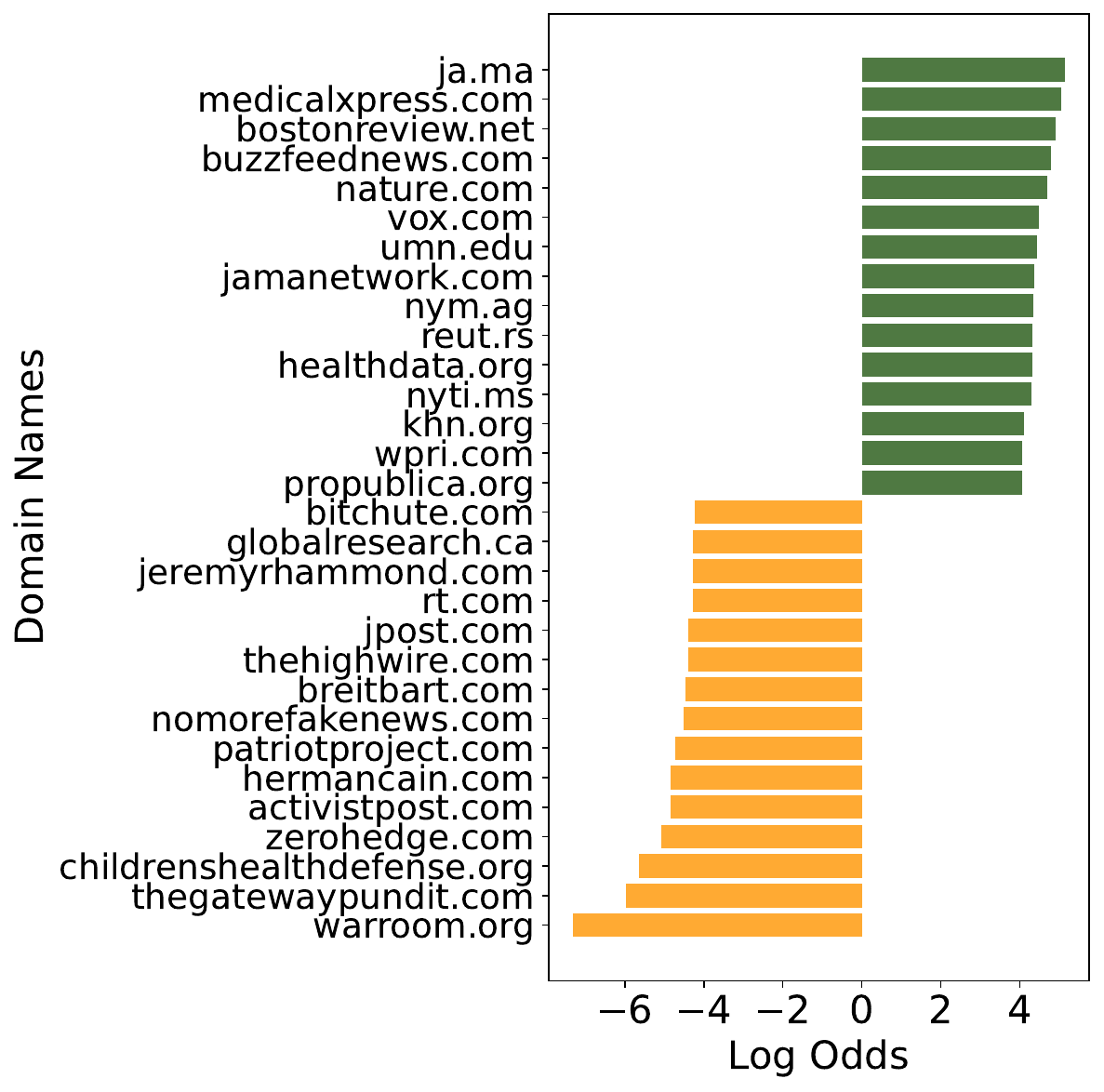}
    \caption{Top sources shared by PHEs and Pseudo Experts} 
    \end{subfigure}
    \begin{subfigure}{0.49\textwidth}
    \includegraphics[width=\textwidth]{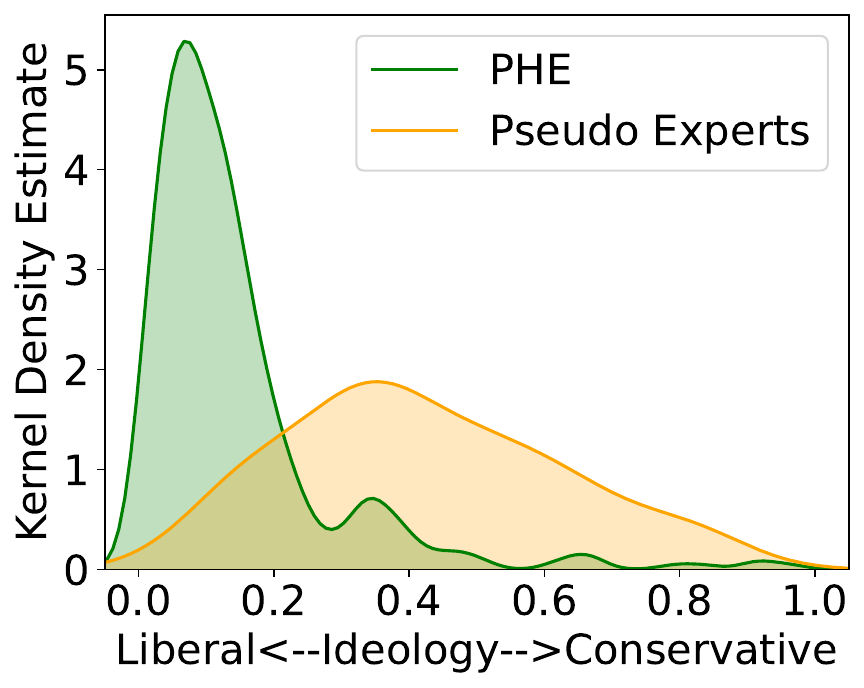}
    \caption{Ideology of sources shared by PHEs and Pseudo Experts} 
    \end{subfigure}
    \caption{Information sources shared by PHEs and pseudo-experts. (a) Top-15 sources that were most likely to be shared by each group. (b) Ideology of sources shared by PHEs and pseudo experts. Mann-Whitney U Test confirms statistically significant differences at $p<0.0001$.}
    \label{fig:ideo_phe_pseudo}
\end{figure}

Further analyzing the content shared by PHEs and pseudo experts, we extract the URLs they shared in their posts and identify the pay-level domains (PLDs) these URLs point to. We compute the log-odds ratio to identify which group is more likely to share each PLD.  Supplementary Figure \ref{fig:ideo_phe_pseudo}(a) shows the top-15 PLDs for both groups. we find that PHEs were more likely to share URLs from highly reputable sources such as Journal of American Medical Association, Nature, Boston Review and New York Times. On the other hand, pseudo experts share more questionable sources such as The Gateway Pundit, Childrens Health Defense, Patriot Project and Russia Today among others. PLDs often have ideological leanings, ranging from liberal ($0$) to least-biased ($0.5$) to conservative ($1$) \cite{MBFC}. 
Supplementary Figure~\ref{fig:ideo_phe_pseudo}(b) compares the distribution of ideological leanings of information sources shared by PHEs and pseudo-experts. While, PHEs tended to share more liberal sources, pseudo experts shared more conservative sources.

\subsection*{Identifying Issue-Relevant Tweets}

We decompose the multi-faceted discussion about the COVID-19 pandemic along various contentious issues: COVID-19 origins, lockdowns and business closures, mask mandates, school closures, therapeutics, healthcare and, vaccines. In order to do so, we rely on methods discussed in \cite{rao2023pandemic,eisenstein2011sparse} to extract issue-relevant keywords from Wikipedia articles. Once we identify keywords, we identify tweets that explicitly mention any of these keywords as being issue-relevant. This approach was validated in \cite{rao2023pandemic} as being able to accurately identify issue-relevant content. Table~\ref{table:sample_of_concern_tweets} illustrates sample tweets from our dataset discussing each issue. 

\begin{table*} [tbh]
    \centering
    \begin{tabular}{l|p{10.5cm}}
        \textbf{Issue} & \textbf{Tweet}\\\hline 
         Education& My take: If you want your child to return to \textcolor{red}{in-person classes} so far NM = only state across the Southern US doing things right [...]\\\hline
         Healthcare& `Our neighbors, our family members': Small-town \textcolor{red}{hospitals} overwhelmed by COVID-19 deaths\\\hline
         Lockdowns& Children's screen time had doubled compared to a year ago. Don’t be alarmed by it yet. Screens allow them to learn and connect with others in times of \textcolor{red}{social distancing} [...]\\\hline
         Masking& Do we need to wear \textcolor{red}{masks} outdoors? I really like the ``2 out of 3" rule by [...] Smart, simple, and clear.  You need 2 of 3: outdoors, masks, distance [...]\\\hline
         Origins& The details of how this \textcolor{red}{virus emerged naturally} are far less exciting than conspiracy theories [...]\\\hline
         Therapeutics& The President's doctors are willing to share specific vital signs today and \textcolor{red}{drug} dosing regimens but claim HIPPA privilege when asked about his chest imaging findings. It’s okay to acknowledge if he has COVID Pneumonia.\\\hline
         Vaccines& CDC plans to launch a new safety program to track COVID \textcolor{red}{vaccines} [...]\\\hline 
    \end{tabular}
    \caption{Issue relevant tweets. Sample tweets for each COVID-19 issue.}
    \label{table:sample_of_concern_tweets}
\end{table*}

We define the origins issue to encompass discussions surrounding the possible causes for the origin of the pandemic, including topics such as pangolins, gain of function research, wet-markets, and bats. The lockdown issue comprises content pertaining to early state and federal mitigation efforts, such as quarantines, stay-at-home orders, business closures, reopening, and calls for social distancing. Discussions related to masking are defined by considerations of face coverings, mask mandates, shortages, and anti-mask sentiment. Education-related content involves tweets regarding school closures, the reopening of educational institutions, homeschooling, and online learning during the pandemic. The healthcare issue deals with conversations on the state of the healthcare system, availability of personal protective equipment, ventilators, oxygen supplies and intensive care units. Discourse around therapeutics encompasses varied alternative treatments proposed to fight COVID-19 infections including Hydroxychloroquine, Ivermectin, plasma therapy, Chinese medicine, colloidal silver and, herbal remedies. The vaccines issue pertains to discussions about COVID-19 vaccines, vaccine mandates, anti-vaccine sentiment, and vaccine hesitancy in the US. 

\subsection*{Identifying Emotions and Morality}
In order to identify emotions expressed in tweets and replies, we employed SpanEmo ~\cite{alhuzali2021spanemo}, a state-of-the-art multi-label emotion detection model. This model was fine-tuned using the SemEval 2018 Task 1e-c dataset~\cite{mohammad2018semeval}. SpanEmo, using a transformer-based architecture. It surpasses previous methods in its ability to capture the correlations among various emotions. When presented with the text of a tweet, the model generates confidence scores for the presence of a wide spectrum of emotions. We later bin these confidence scores using a $0.5$ threshold to binarize the output. The emotions it can identify include \textit{anticipation}, \textit{joy}, \textit{love}, \textit{optimism}, \textit{anger}, \textit{disgust}, \textit{fear}, \textit{sadness} and, \textit{pessimism}. The definitions for these emotions are borrowed from \cite{mohammad2018semeval} as follows:

\begin{itemize}
    \item anticipation (also includes interest, vigilance)
    \item joy (also includes serenity, ecstasy)
    \item love (also includes affection)
    \item optimism (also includes hopefulness, confidence)
    \item anger (also includes annoyance, rage)
    \item disgust (also includes disinterest, dislike, loathing)
    \item fear (also includes apprehension, anxiety, terror)
    \item sadness (also includes pensiveness, grief)
    \item pessimism (also includes cynicism, no confidence)
\end{itemize}


Prior research has shown that emotional and moral language in social media messages impacts how they are received by the audiences and engagement~\cite{brady2021social,kramer2014experimental}. The Moral Foundations Theory (MFT) ~\cite{haidt2007morality}  provides a framework for understanding how moral values shape people's political attitudes and behaviors. MFT proposes that individuals' values and judgments can be described by five moral virtue/vice pairs - care/harm, fairness/cheating, loyalty/betrayal, authority/subversion, and sanctity/degradation. More specifically,


\begin{itemize}
    \item Care/harm: This foundation revolves around the concept of empathy and compassion. People who prioritize this foundation value caring for others and preventing harm. They are sensitive to the suffering of others and strive to promote their well-being.
    \item Fairness/cheating: This foundation is concerned with issues of justice, reciprocity, and fairness. Individuals who emphasize this foundation are attuned to issues of equality, fairness, and proportionality. They believe in treating others fairly and oppose exploitation and unfair advantage.
    \item Loyalty/betrayal: People who prioritize loyalty value group cohesion, allegiance, and solidarity. They are inclined to support and defend their in-groups, whether it be family, community, or nation, and perceive betrayal or disloyalty as morally reprehensible.
    \item Authority/subversion: This foundation centers on respect for authority, tradition, and hierarchy. Individuals who emphasize this foundation value social order, respect for authority figures, and obedience to legitimate institutions and norms. They believe that maintaining authority and order is essential for a stable society.
    \item Sanctity/degradation: This foundation involves the reverence for purity, sanctity, and sacredness. People who prioritize this foundation are concerned with issues related to cleanliness, moral purity, and spiritual transcendence. They may view certain actions, objects, or behaviors as inherently sacred or profane.
\end{itemize}

Our morality detection model is trained on the BERT transformer-based pretrained language model by \cite{devlin2018bert}. The training process involves three Twitter datasets, a manually annotated COVID dataset \cite{rojecki2021moral}, the Moral Foundation Twitter Corpus dataset covering six different topics \cite{Hoover2020moral}, and a dataset of political tweets from US congress members \cite{johnson-goldwasser-2018-classification}. By incorporating an in-domain training set focused on COVID-19, along with other diverse datasets spanning various topics, we enhance the model's generalizability for application to target data as discussed in \cite{guo2023data}.



\begin{figure}[htb]
    \centering
    \includegraphics[width=0.8\textwidth]{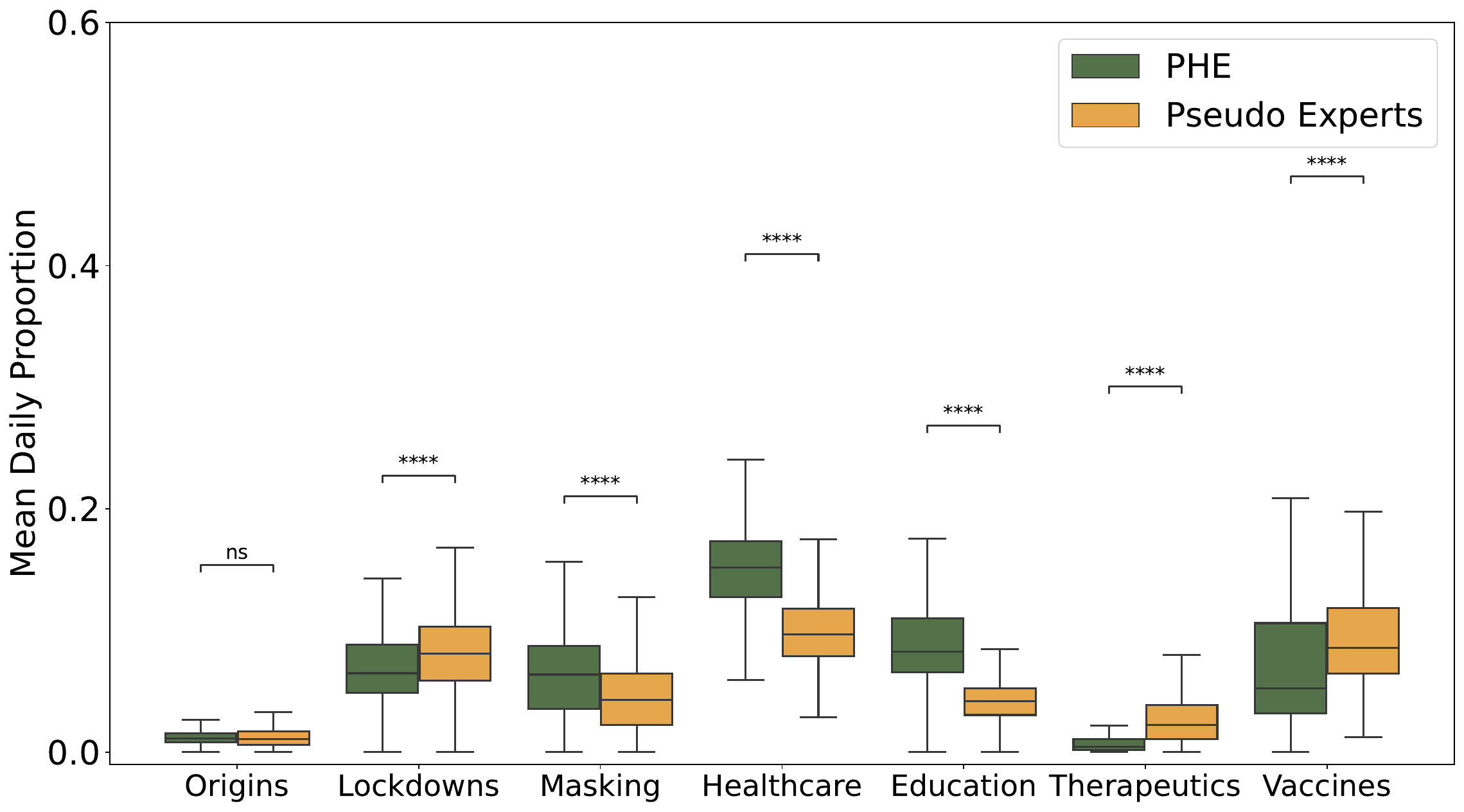}
    \caption{Comparing the activity of PHEs and pseudo-experts along each issue. Box plots compare the daily proportion of issue related tweets from PHEs and pseudo-experts. Mann-Whitney U Test with Bonferroni correction is used to assess significance. * indicates significance at $p<0.05$, ** - $p<0.01$, *** - $p<0.001$, **** - $p<0.0001$ and, ns - not-significant.}
    \label{fig:concerns_compare}
\end{figure}
\section*{Results}

\subsection*{Messaging about COVID-19 Issues}

More than half the tweets from PHEs and pseudo-experts mention at least one of the seven COVID-19 issues we identified. 
Figure \ref{fig:concerns_compare} compares the average daily share of tweets from both groups on each issue.  Overall, we find that pseudo-experts tend to be more vocal on the issues of lockdowns, therapeutics and vaccines while, PHEs generate more content about masking, healthcare and education. We do not witness any significant differences in the discourse about origins of the virus. 
These trends reflect the attention to issues by each group prior to President Biden's inauguration, which is the period covered by this study.

To better summarize the varied perspectives expressed by PHEs and pseudo experts on the seven issues of interest, we randomly sample 25 tweets for the two groups across these issues and prompt ChatGPT to provide the broad perspective being expressed using the prompt:
\begin{quote}
    \texttt{Summarize the perspectives being expressed about <Issue> in these tweets: <T>}
\end{quote}
where \texttt{<Issue>} is one of \texttt{[Origins, Lockdowns, Masking, Education, Healthcare, Therapeutics, Vaccines]} and \texttt{<T>} represents a concatenation of the 25 tweets that were randomly sampled for each issue and group pair. 

The results presented in Supplementary Table \ref{tab:perspec} demonstrate the contrasting viewpoints between the two groups on various issues. Regarding the origins of the virus, PHEs generally lean towards the belief that it originated in a laboratory, albeit with some skepticism, while pseudo-experts heavily criticize China and its potential involvement in gain of function research. PHEs emphasize the importance of ongoing vigilance, adherence to stay-at-home orders, and widespread use of masks, whereas pseudo-experts question the effectiveness of lockdowns and mask mandates and criticize government intervention in these areas. On the topic of therapeutics, PHEs urge caution against self-prescribing drugs like Hydroxychloroquine, Azithromycin, and Ivermectin without evidence of their efficacy in treating COVID-19, whereas pseudo-experts advocate for the use of these medications.

\begin{figure}[htb]
    \includegraphics[width=\textwidth]{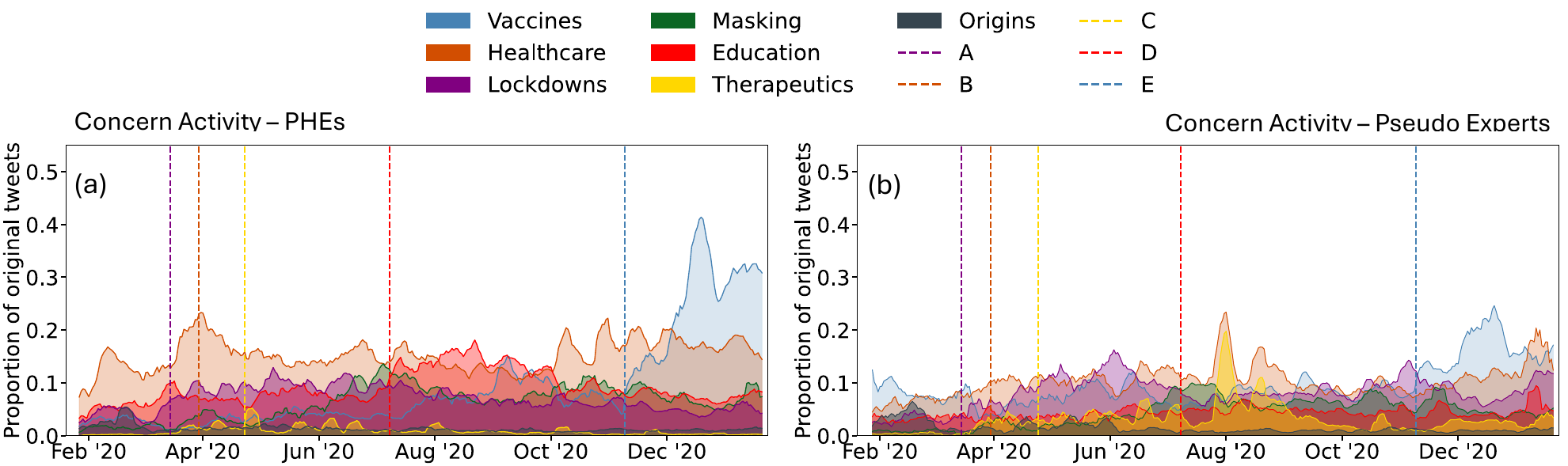}
    \caption{Timeline of attention to issues. Daily fraction of original tweets posted by (a) PHEs and (b) pseudo-experts related to each issue. We use 7-day rolling average to reduce noise. Major events are marked by vertical lines (A-G). (A) Lockdowns: March 15, 2020, stay-at-home orders start being issues across the mainland United States; (B) Healthcare: March 30, 2020; (C) Therapeutics: April 24, 2020 as President Trump proposes fighting off the virus with bleach; (D) Education: July 8, 2020,  President Trump calls for schools to reopen; (E) Vaccines: November 9, 2020, Pfizer reports 93\% efficacy in Phase-3 trials.}
    \label{fig:phe_concerns}
\end{figure}

Next, we look at the temporal patterns of issue-related discussions. Figure \ref{fig:phe_concerns}(a) and (b) shows the daily share of posts from each group about the issues. When stay-at-home orders start being issued in mid-March 2020, we see a rise in lockdown related discussions from PHEs. Lockdowns-related discourse from pseudo experts picked up steam mid-April amidst the calls for reopening the economy and further accelerated in early June 2020  during Black Lives Matter protests, when they were critical of mass protests. As COVID-19 cases rose in late March 2020, we see a spike in healthcare related discourse from PHEs with growing calls for emergency preparedness in terms of improving access to personal protective equipment and ventilators. We do not observe a corresponding increased from pseudo-experts. We see a small spike in therapeutics related discussions among PHEs following President Trump's April 24, 2020 comment on using bleach to ward off the COVID-19 virus. Almost immediately following the Federal Drug Administration (FDA)'s issuance of a Emergency Use Authorization on various therapeutics such as Hydroxychloroquine (HCQ) on March 28, 2020, we see an immediate increase in therapeutics related discussions from pseudo-experts. However, we see highest share of posts from them on July 26, 2020 when then-White House Chief of Staff Mark Meadows announced that alternative therapeutics would be coming soon. We also see spikes in education related discussions from PHEs and pseudo experts following President Trump's July 8, 2020 call to reopen educational institutions. However, the spikes were for very different reasons - PHEs expressed increased skepticism towards reopening schools while pseudo-experts supported the reopening call. The largest spikes for both groups are for vaccine-related discussions following Pfizer's announcement of successful COVID-19 Phase-3 vaccine trials (10\% to 37\% for PHEs and 20\% to 32\% for pseudo-experts). 

\begin{figure}[!htb]
    \centering
    \includegraphics[width=0.75\linewidth]{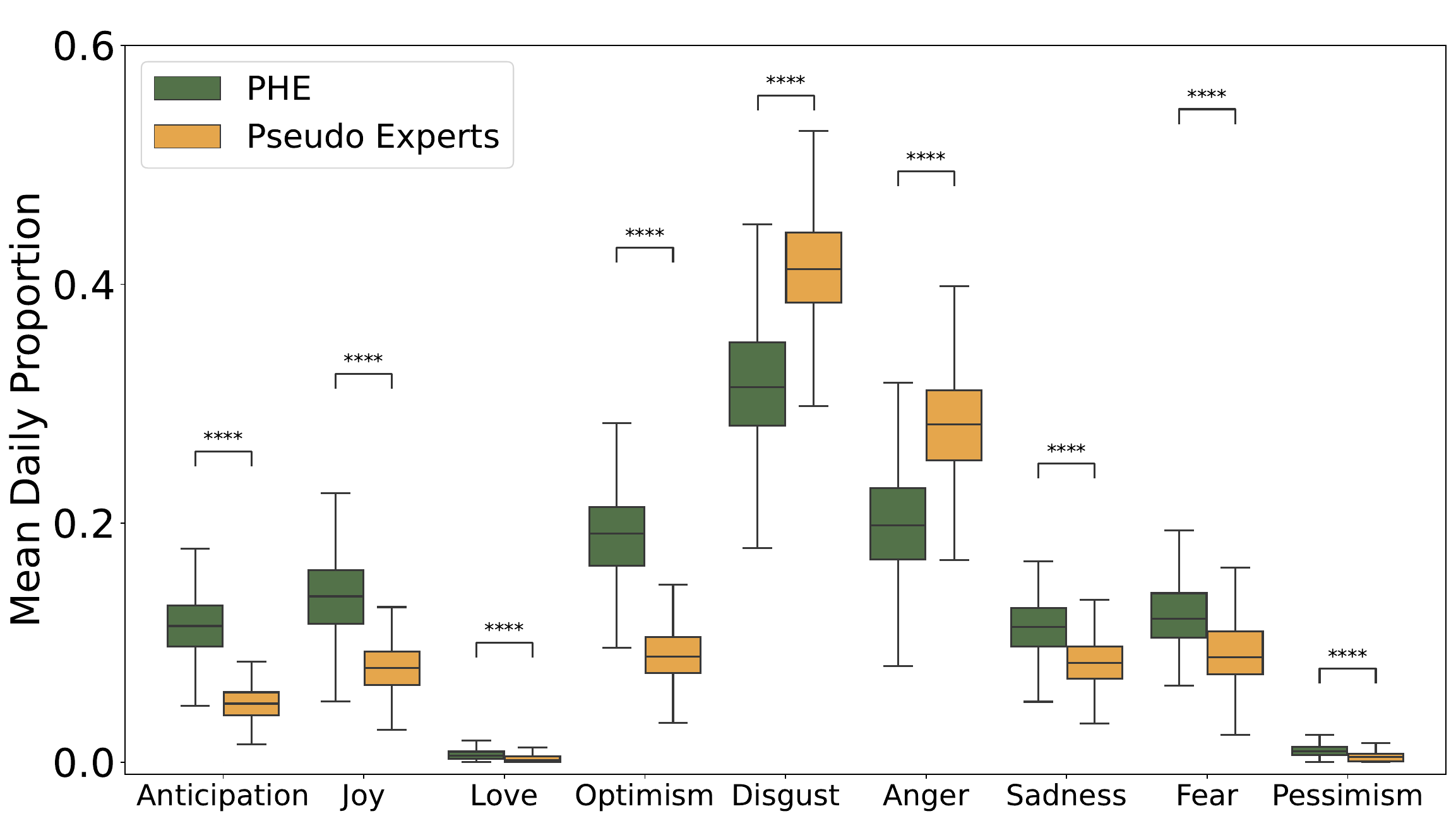}
    \caption{Comparing emotions expressed by PHEs and pseudo-experts in messages about the issues. Box plots compare the daily proportion of tweets from PHEs and pseudo-experts expressing various emotions. Mann-Whitney U Test with Bonferroni correction is used to assess significance. * indicates significance at $p<0.05$, ** - $p<0.01$, *** - $p<0.001$, **** - $p<0.0001$ and, ns - not-significant.}
    \label{fig:emotion_comparison}
\end{figure}
\begin{figure}[!htb]
    \centering
    \includegraphics[width=0.9\textwidth]{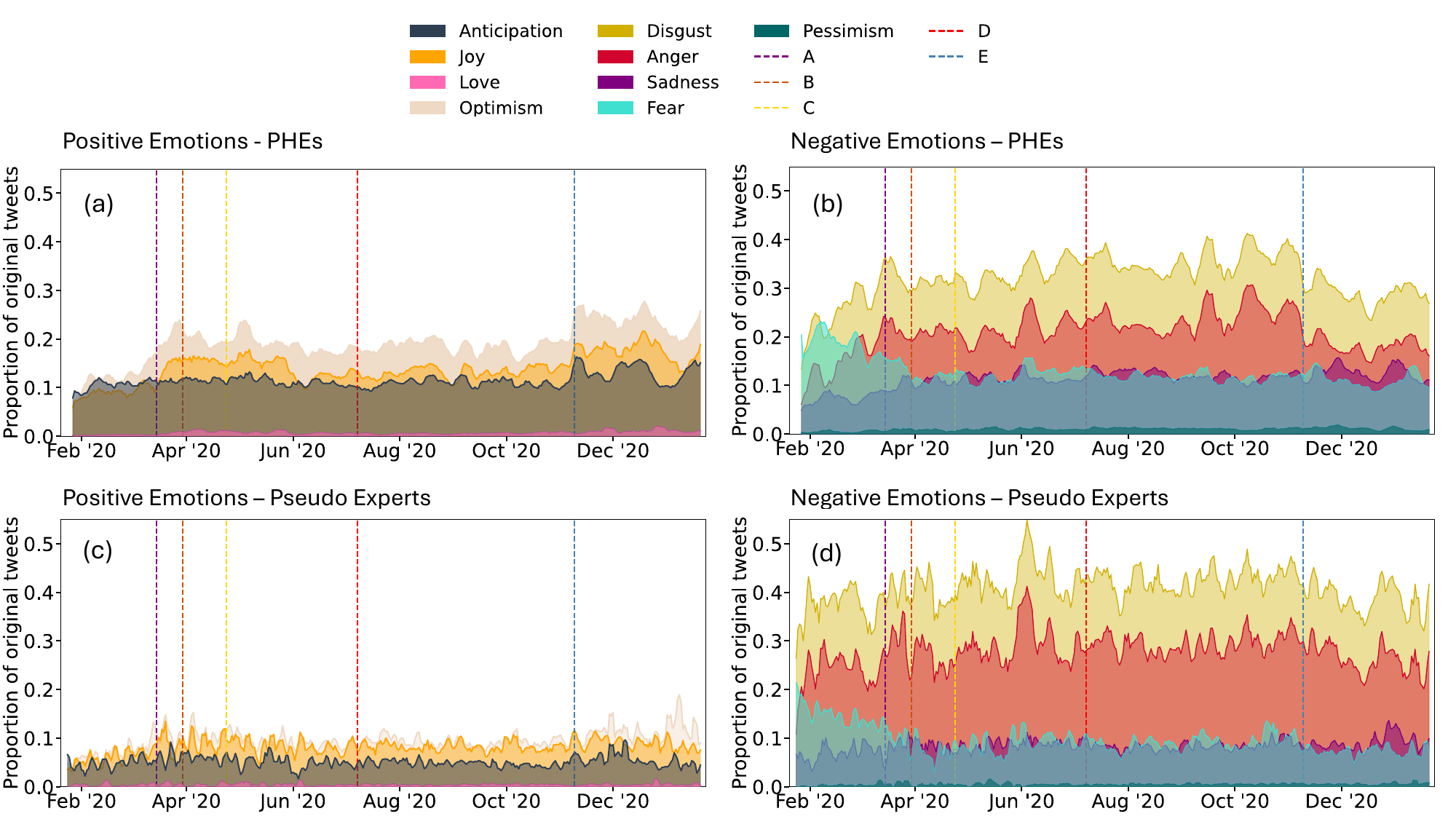}
    \caption{Dynamics of emotions. Daily fraction of tweets from PHEs and pseudo-experts  expressing (a-b) positive (optimism, joy, anticipation and love) and (c-d) negative (disgust, anger, sadness, fear and pessimism) emotions. We use 7-day rolling average to reduce noise. Major events are marked with vertical lines (A-G). (A) Lockdowns: March 15, 2020, stay-at-home orders start being issues across the mainland United States; (B) Healthcare: March 30, 2020; (C) Therapeutics: April 24, 2020 as President Trump proposes fighting off the virus with bleach; (D) Education: July 8, 2020,  President Trump calls for schools to reopen; (E) Vaccines: November 9, 2020, Pfizer reports 93\% efficacy in Phase-3 trials.}
    \label{fig:phe_emotions}
\end{figure}

\subsection*{Emotional and Moral Language}

Figure~\ref{fig:emotion_comparison} compares the distribution of the daily fraction of tweets posted by PHEs and pseudo-experts expressing a certain emotion. Overall, PHEs express more positive emotions such as anticipation, joy and optimism and also more low arousal negative emotions like sadness and fear, whereas pseudo-experts express more high arousal negative emotions anger and disgust. Interestingly, we do not see much love or pessimism in our data. 

\paragraph*{Dynamics of Affect}
Emotions fluctuate over time and in response to events. Figure \ref{fig:phe_emotions} illustrates the temporal dynamics of positive emotions expressed by PHEs and pseudo-experts. We leverage ChatGPT to summarize changes in emotions expressed and are discussed further in Supplementary Table \ref{tab:emotions_summ}. Optimism and joy among PHEs surge following the announcement of stay-at-home orders post March 15, 2020. This can be attributed to  factors such as gratitude for guidance by then-New York Governor Andrew Cuomo and enhanced accessibility to COVID-19 testing. Similarly, we note a corresponding albeit smaller increase among pseudo-experts, particularly in response to President Trump's management of the pandemic and France's endorsement of HCQ as a viable COVID-19 treatment.

Another surge in joy, anticipation, and optimism among PHEs occurs after November 9, 2020, following Pfizer's announcement of successful Phase-3 trials for its COVID-19 vaccine. PHEs hailed this development as a remarkable achievement and anticipated emergency use authorization  from the FDA. While positive emotions also increased among pseudo-experts, the magnitude was notably lower. Pseudo-experts expressed optimism regarding the success of Operation Warp Speed, the imminent reopening of businesses, and the introduction of Lilly's monoclonal antibody drug. 

Negative emotions such as disgust and anger escalated for both groups post March 15, 2020, with a more pronounced increase among pseudo-experts. The upsurge in anger and disgust within each group stemmed from different reasons. PHEs expressed disappointment with the measures taken by the Trump administration to combat the virus, whereas pseudo-experts voiced skepticism concerning the World Health Organization's interactions with China, Governor Cuomo's management of public transportation in New York, and the effectiveness of lockdowns in containing COVID-19. Although both groups experienced parallel declines in anger and disgust after US elections on November 9, 2020, the reductions were more significant among PHEs. 



\begin{figure}[!ht]
    \centering
    \includegraphics[width=0.97\linewidth]{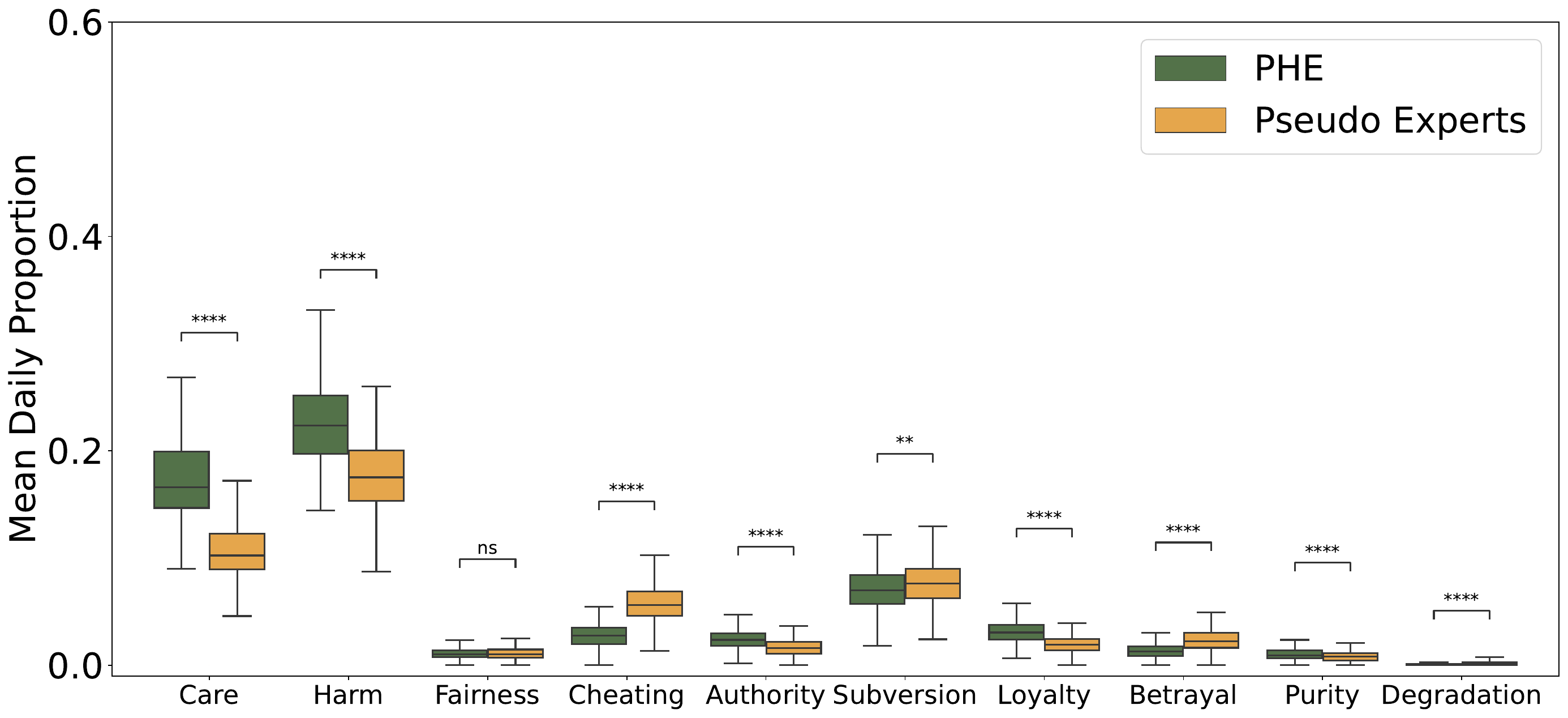}
    \caption{Moral comparison. Comparing moral attitudes expressed overall by PHEs and pseudo-experts along each issue. Box plots compare the daily proportion of tweets from PHEs and pseudo-experts expressing various moral foundations. Mann-Whitney U Test with Bonferroni correction is used to assess significance. * indicates significance at $p<0.05$, ** - $p<0.01$, *** - $p<0.001$, **** - $p<0.0001$ and, ns - not-significant.}
    \label{fig:mft_comparison}
\end{figure}

\begin{figure}[!ht]
    \centering
    \includegraphics[width=\textwidth]{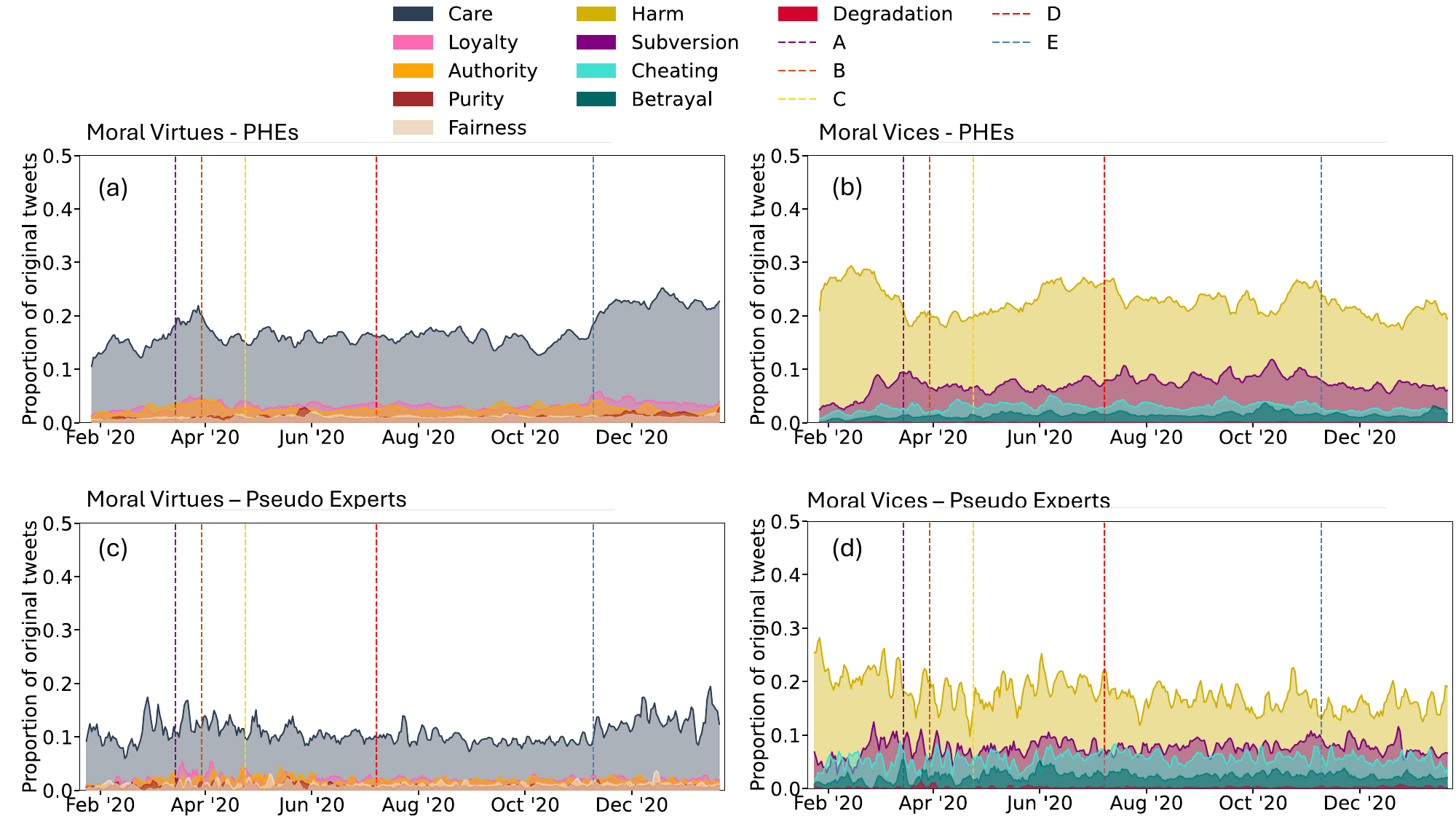}
    \caption{Dynamics of moral attitudes. Daily fraction of tweets from PHEs (a,c) and pseudo-experts (b,c) expressing positive (care, fairness, loyalty, authority and purity)  and negative (harm, cheating, betrayal, subversion and degradation) moral foundations. We use 7-day rolling average to reduce noise. Major events are marked with vertical lines (A-G). (A) Lockdowns: March 15, 2020, stay-at-home orders start being issues across the mainland United States; (B) Healthcare: March 30, 2020; (C) Therapeutics: April 24, 2020 as President Trump proposes fighting off the virus with bleach; (D) Education: July 8, 2020,  President Trump calls for schools to reopen; (E) Vaccines: November 9, 2020, Pfizer reports 93\% efficacy in Phase-3 trials; (F) Masking: May 14, 2021, CDC sets aside indoor masking requirements; (G) Origins: May 23, 2021, Wall Street Journal reports a low confidence assessment that lab leak was the most likely cause for COVID-19.}
    \label{fig:phe_mfts}
\end{figure}

We also examine the use of moral language by the two groups. Supplementary Figure \ref{fig:mft_comparison} compares the distribution of the daily share of tweets expressing each moral foundation. Overall, PHEs use more  positive moral language, emphasizing the dimensions of care, fairness, authority, loyalty and purity, while  pseudo-experts tend to prefer the negative moral dimensions of harm, cheating, subversion and betrayal.
The differences in use of moral language are more subdued compared to those for emotions. Supplementary Figure~\ref{fig:phe_mfts}  illustrates the temporal dynamics of positive and negative moral language used by PHEs and pseudo-experts. We summarize the positive spikes using ChatGPT in Supplementary Table \ref{tab:mft_summ}.


We witness an increase in the expression of care from PHEs post stay at home orders. This increase is marked by calls for widespread lockdown measures, testing and relief proposals for low-income households. However, there is a marginal decline in care language from pseudo-experts. Harm moral language decreases for both groups but more so for PHEs. In response to Pfizer's successful phase-3 trials, the use of care language increases for both PHEs and pseudo-experts coupled with a decline in the use of harm language. Both groups express care in discussing how the introduction of vaccines could bring an end to the pandemic. Additionally, pseudo-experts express concerns about the safety of the mRNA vaccines and criticize Bill Gates' call for vaccine mandates. 


\paragraph*{Asymmetries in Emotions and Moral Language}

Public health experts and pseudo-experts had conflicting priorities.  PHEs promoted vaccination and advocated for stringent non-pharmaceutical interventions to curb the spread of the virus. In contrast, pseudo-experts expressed skepticism towards such interventions, emphasizing personal choice. We examine how these differences were manifested in the emotional and moral language used by the two groups.

\begin{figure}[!htb]
    \centering
    \includegraphics[width=\textwidth]{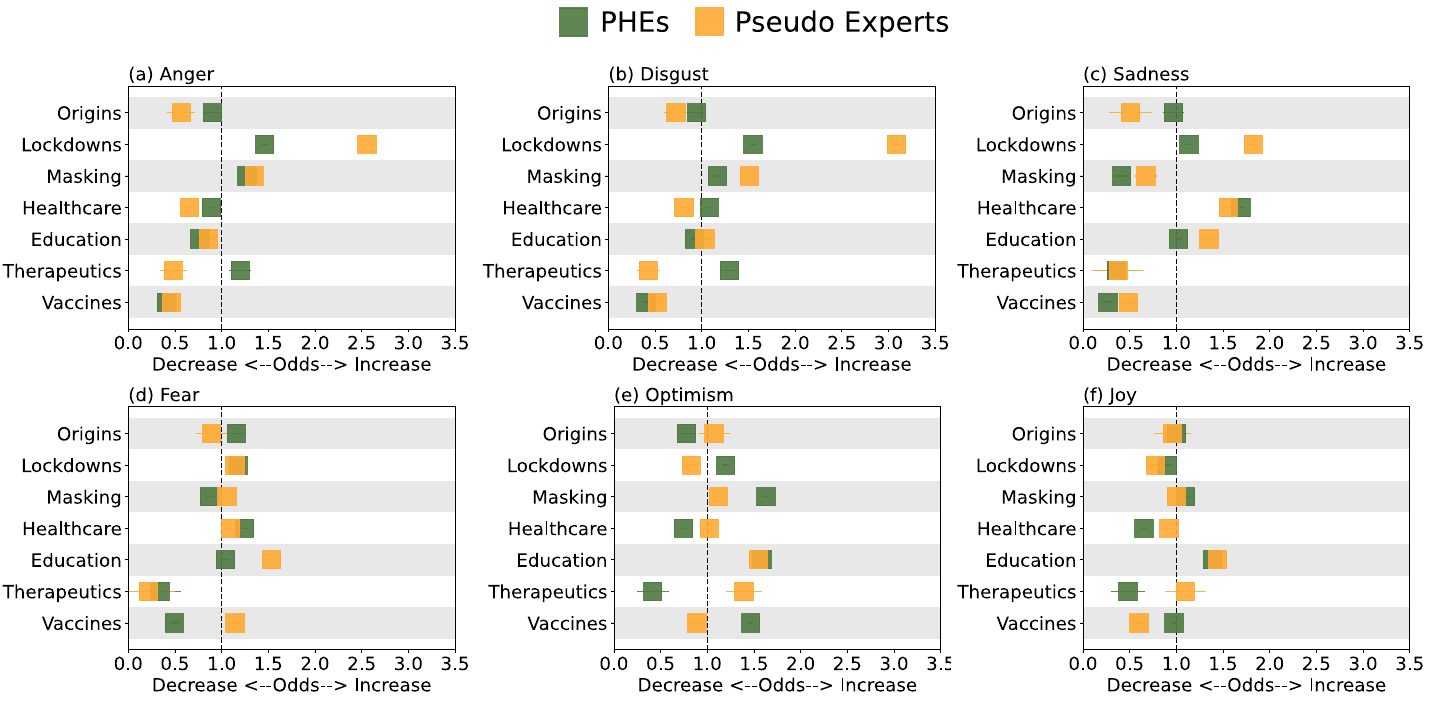}
    \caption{Comparing emotions conveyed by PHEs and pseudo-experts across various topics. Displays the odds ratio of a tweet's relevance to a particular issue based on the expression of specific emotions by PHEs. An odds ratio exceeding 1 suggests that when a particular issue is discussed, there are increased odds of the tweet expressing an emotion compared to when the issue isn't ; an odds ratio of 1 indicates equal odds, while a ratio below 1 signifies lower odds.}
    \label{fig:phe_vs_pseudo_emotions}
\end{figure}

To quantify issue-specific variation in emotions and moral language use by the two groups, we conduct a multivariate logistic regression analysis at the tweet level for each emotion and moral foundation. We examine the relationship between the issue discussed (independent variable) and the emotions or moral foundations expressed (dependent variables). Additionally, the model incorporates a categorical variable to delineate between different groups. To account for potential differences in the emotional responses of the two groups, we introduce an interaction term between issues discussed and the group variable. We formulate the model separately for each issue as follows:

$$
\verb|<Emotion>| \sim origins + lockdowns + masking + education + healthcare + therapeutics + vaccines + \\
$$
$$
(origins + lockdowns + masking + education + healthcare + therapeutics + vaccines) * group,
$$
where, \verb|group| distinguishes between PHEs and pseudo-experts. We run separate regression models for each emotion. 
The coefficients for the main effects represent the change in the log-odds of the emotion for the PHEs when discussing an issue while holding all other issues constant. On the other hand, the sum of coefficients of the main effects and interaction effects, quantify the change in log-odds for the pseudo-experts. For example, if the coefficient for lockdowns is positive, it suggests that when lockdowns are being discussed there is an increase in the expression of a particular emotion from when they do not discuss lockdowns. 


\begin{figure}[!htb]
    \centering
    \includegraphics[width=\textwidth]{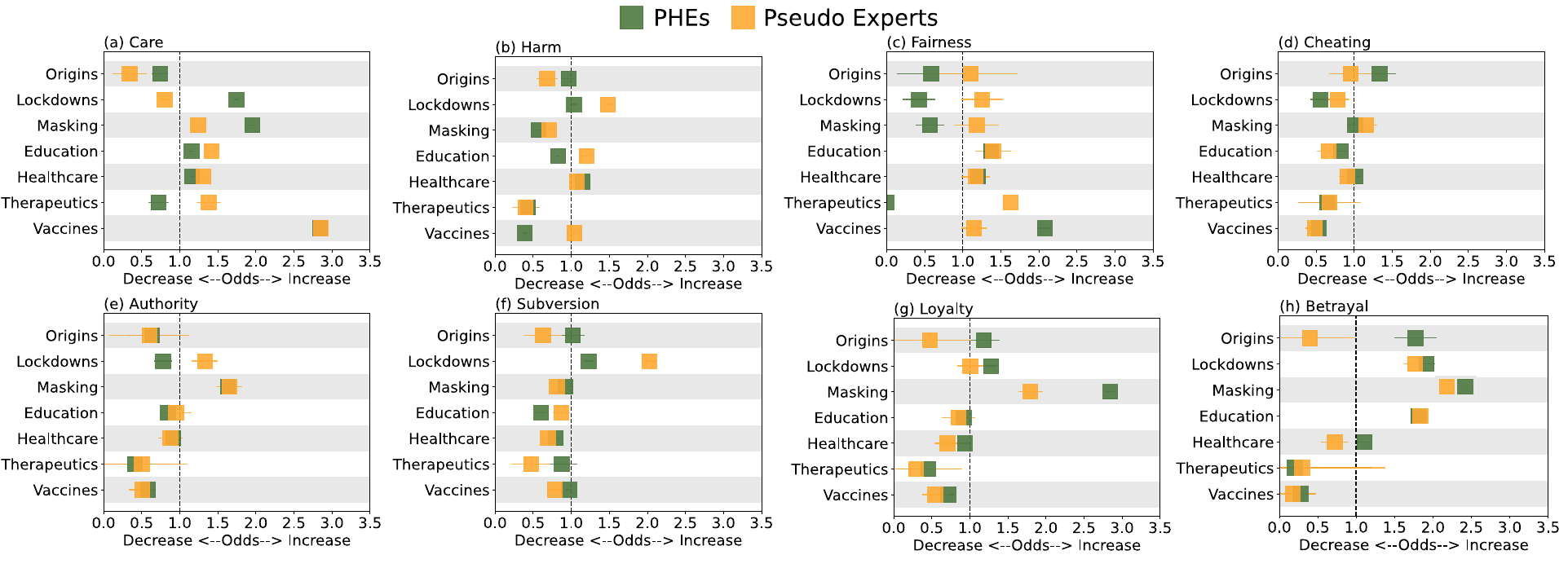}
    \caption{Comparing moral foundations conveyed by PHEs and pseudo-experts across various topics. Displays the odds ratio of a tweet's relevance to a particular issue based on the expression of specific moral foundations by PHEs. An odds ratio exceeding 1 suggests that when a particular issue is discussed, there are increased odds of the tweet expressing a moral foundation compared to when the issue isn't being discussed; an odds ratio of 1 indicates equal odds, while a ratio below 1 signifies lower odds.}
    \label{fig:phe_vs_pseudo_mft}
\end{figure}

Figure \ref{fig:phe_vs_pseudo_emotions} compares the log-odds along with the corresponding standard errors of estimation, to show which group used more emotional language to frame a specific issue. The plot highlights differences between the two groups and gives insights into emotionally charged issues. The biggest gap in emotions appears on the issue of lockdowns, where pseudo-experts are far more likely to express anger, disgust and sadness than PHEs. This position is consistent with the efforts to end the lockdowns (see Great Barrington declaration~\cite{burki2021herd}). The second largest gap in emotions appears in the discussion of therapeutics, where PHEs are more likely to express anger and disgust, but pseudo-experts are less likely to use these emotions. Pseudo-experts also use more positive language with more joy and optimism when talking about therapeutics, in contrast to PHEs, consistent with the highly contentious debates about this issue.
Other notable differences include pseudo-experts expressing more fear and less joy about vaccines, while PHEs express less fear and more optimism. 


We carry out a similar analysis of the moral language used by the two groups of users. 
%
%
The coefficients for the main effects represent the change in the log-odds of the moral foundation for the PHEs when an issue is being discussed while holding all other issues constant and the sum of coefficients of the main effects and interaction effects quantifies the change in log-odds for the pseudo-experts. Figure \ref{fig:phe_vs_pseudo_mft} compares the log-odds, along with the corresponding standard errors of estimation, to illustrate which group relied more heavily on a given moral foundation when framing a specific issue. 

The differences in moral language usage are less pronounced compared to emotions. PHEs tend to emphasize care and loyalty in discussions of lockdowns and masking, consistent with their use of prosocial messaging focusing on the collective benefits of these measures. Conversely, pseudo-experts tend to convey more notions of harm, fairness, authority, and  subversion when addressing lockdowns. This is in line with this issue being extremely contentious for them. Surprisingly, pseudo-experts are more likely to use fairness to frame their discussions of all issues, except vaccines. 


The comparison of emotions and moral foundations between PHEs and pseudo-experts, makes evident their conflicting positions on key issues pandemic related issues, notably through the higher use of negative emotions on issues central to the other group. PHEs tend to focus on discussions related to vaccines, healthcare, and education and these are issues on which we see more negative emotional framing by the pseudo-experts. In contrast, pseudo-experts are more focused on therapeutics and alternative treatments. Similarly, their negative framing of lockdowns and vaccines reflects their disapproval of the issues that were heavily promoted by PHEs. 
This divergence of affect underscores the  polarization in the society at large. Understanding these differences is crucial for informing public health communication efforts that promote consensus within different segments of the population. 


\subsection*{Affective Polarization in Health Communication}

Studies show that public response to the pandemic became polarized fairly quickly, with political partisanship shaping online activity and discussions about the pandemic already in the early stages of the pandemic~\cite{jiang2020political,rao2021political}. Online discussions also became emotionally polarized: when interacting with members of the opposite party, Twitter users expressed more anger and disgust, more toxicity, and less joy than in their interactions with same-party members~\cite{feldman2023affective}. Such interactions are characteristic of affective polarization~\cite{iyengar2015fear}, patterns of in-group love and out-group hate  that have contributed to the growing partisan divide and the erosion of trust between the two parties in US. As a result, partisanship predicted the adoption of pandemic prevention measures more than other factors~\cite{grossman2020political}.

\begin{figure}
    \centering
    \includegraphics[width=0.9\textwidth]{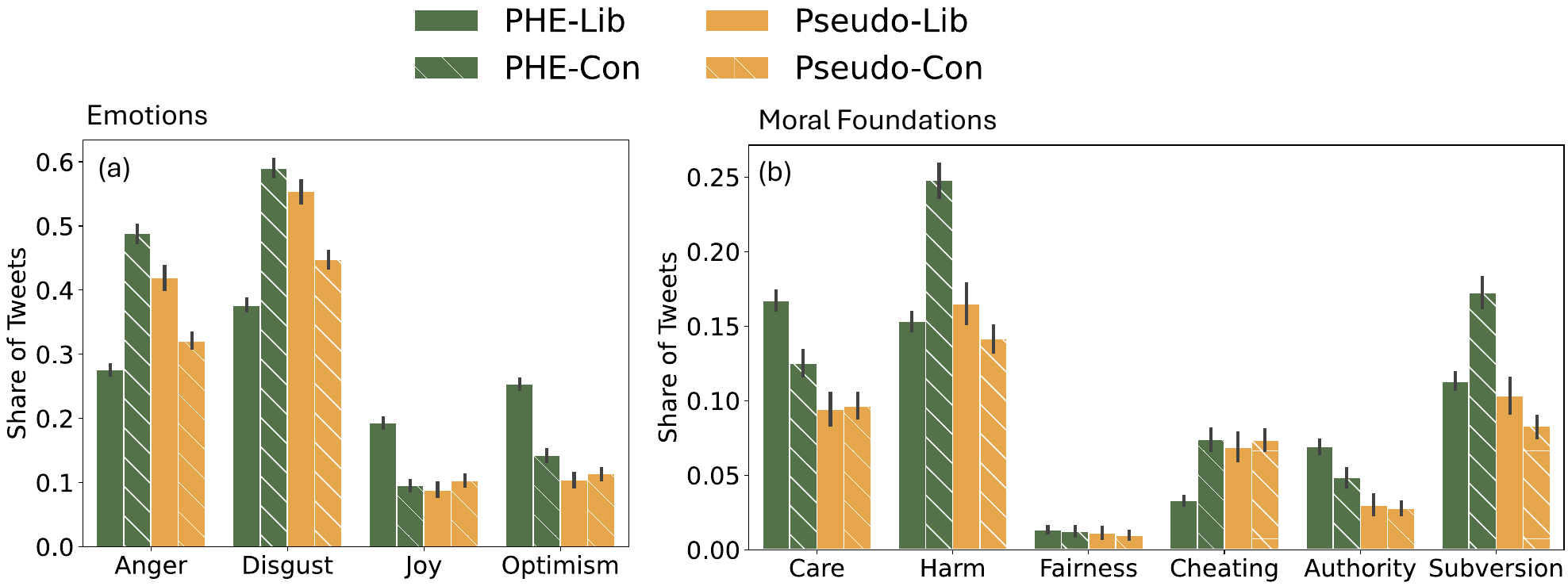}
    \caption{Asymmetries in affect towards elites. (a) PHEs direct more negative emotions (anger, disgust) toward conservative elites  and more positive emotions (joy, optimism) in their mentions of liberal elites, which is a hallmark of affective polarization. In contrast pseudo-experts direct more negativity toward liberal elites. With respect to moral language (b), express more subversion in their mentions of conservative elites, in contrast to pseudo-experts.
}
\label{fig:lib_con_compare}
\end{figure}

To measure affective polarization, we analyze the emotional language PHEs and pseudo-experts directed at the political elites in their original tweets. 
We use a previously curated  list of Twitter handles of over 1.7K political elites,\footnote{\url{https://github.com/sdmccabe/new-tweetscores}\label{ideo_scores}} which include current and former Senators, Representatives, and media pundits.
Figure~\ref{fig:lib_con_compare} shows the proportion of tweets from each group with mentions of political elites that express various affect. For instance, PHE-Lib indicates the share of PHEs' tweets with a greater frequency of references to liberal elites compared to conservative ones, expressing a specific emotion or moral foundation.

The figure reveals that PHEs post as liberal partisans: when they mention conservative elites in their tweets, they use more negative emotions and moral subversion, but when they mention liberal elites, they express more positive emotions. In contrast, pseudo-experts are conservative partisans: they direct more negativity toward liberal elites while expressing more positivity towards other conservative elites.

\begin{figure}[!ht]
    \centering
    \includegraphics[width=\textwidth]{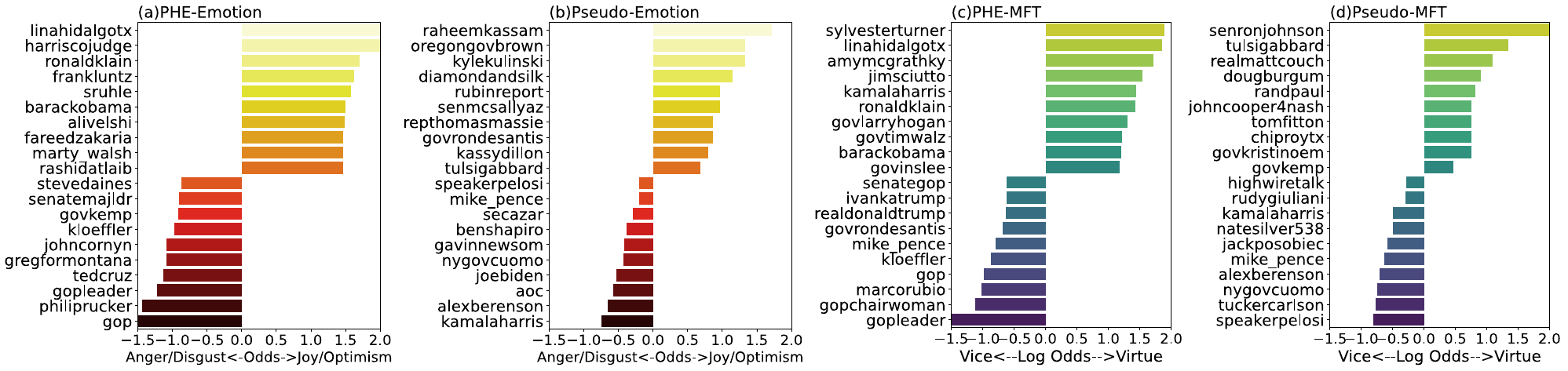}
    \caption{Engagement with Political Elites. Top-10 political elites more likely to be referenced by PHEs and pseudo-experts when expressing positive vs negative emotions and moral foundations. (a)top-10 accounts when PHEs express anger or disgust vs joy or optimism. (b) top-10 accounts when pseudo-experts express anger or disgust vs joy or optimism. (c) top-10 accounts when PHEs express moral virtues (care, fairness, authority and loyalty) vs moral vices (harm, cheating, subversion and betrayal), (d)top-10 accounts when pseudo-experts express moral virtues vs moral vices.}
    \label{fig:pol_odds}
\end{figure}

\begin{figure}[!ht]
    \centering
    \includegraphics[width=\textwidth]{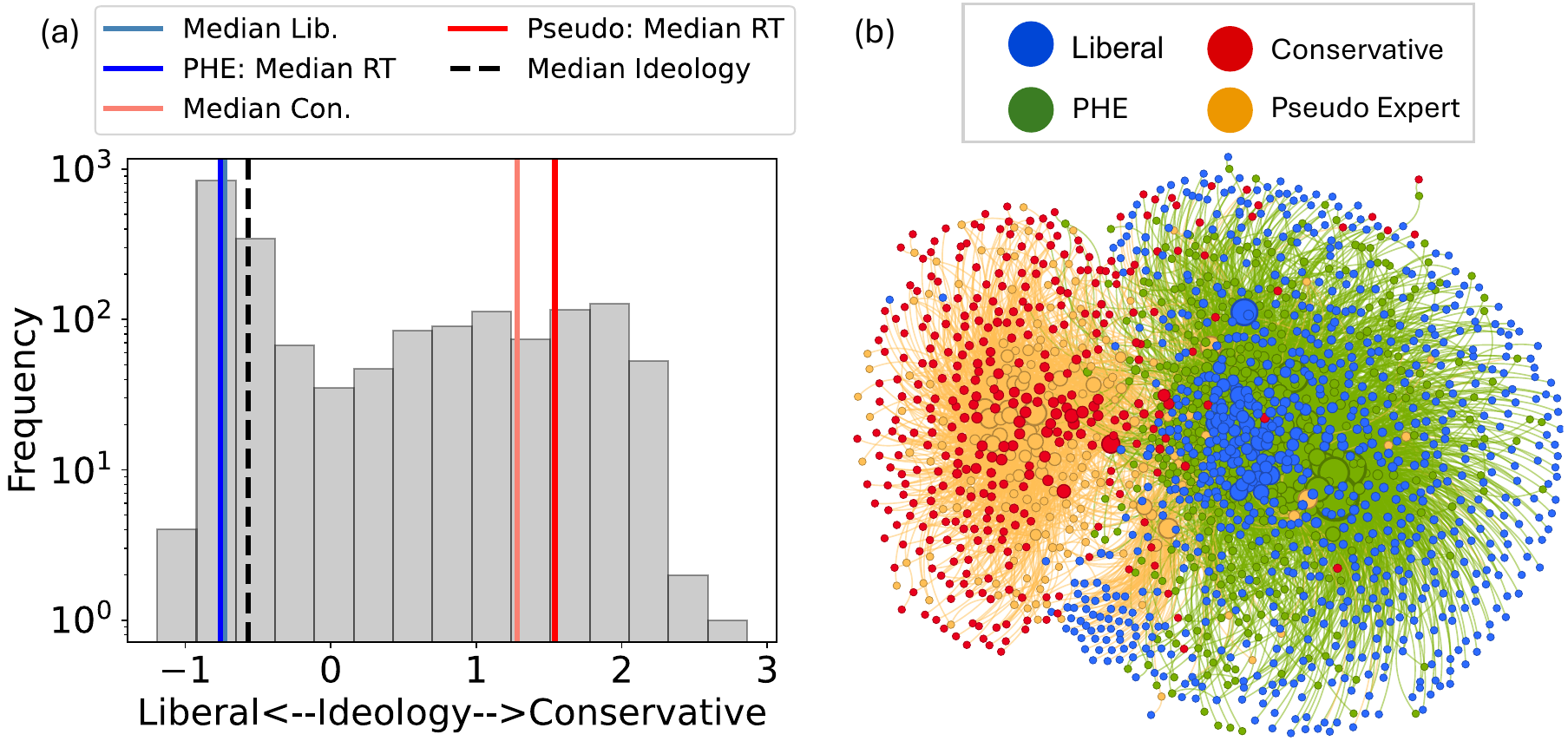}
    \caption{Retweet Interactions with Political Elites. Retweet interactions of Public Health Experts and Pseudo Experts with political elites.(a) shows the distribution of ideology scores of political elites. The median ideology score of the elites retweeted by PHEs are more liberal than the median liberal elite. Similarly, the median score of the elites retweeted by pseudo-experts is more conservative than the median conservative elite. (b) shows the ideological clustering in retweet preferences of PHEs and pseudo-experts. PHEs prefer to retweet liberal elites while pseudo-experts retweet conservative elites.}
    \label{fig:ideo_rt}
\end{figure}

Figure~\ref{fig:pol_odds} lists the top ten accounts that are more likely to be mentioned by PHEs and pseudo-experts positively or negatively. 
We then assess the retweet interactions of public health and pseudo-experts with political elites. To this end, we construct a bipartite network comprising of directed edges from PHEs/pseudo-experts to political elites with each edge indicating the political elite retweeted by a PHE or pseudo-expert. Supplementary Figure \ref{fig:ideo_rt}(a) shows the distribution of ideology estimates for political elites active on Twitter during COVID-19. The ideology estimates are were obtained from \footref{ideo_scores} and were calculated using methods described in \cite{barbera2015birds}. The median ideology score of political elites on Twitter is $-0.56$ which is consistent with the past reports identifying the liberal skew of Twitter \cite{pew2019twitter}. The median liberal and conservative elite have an ideology score of $-0.73$ and  $1.28$ respectively. Interestingly, the median ideology score of the elites retweeted by PHEs and pseudo-experts are $-0.76$ and $1.54$ indicating that a considerable share of the elites retweeted by PHEs and pseudo-experts are more liberal and more conservative respectively, than the median liberal and conservative elite. In Supplementary Figure \ref{fig:ideo_rt}(b) we visualize this network to find two highly clustered interactions. The color of the edge indicates the color of the source node i.e., PHEs (green) or pseudo-experts (orange). We find that PHEs mostly retweet liberal elites whereas, pseudo-experts mostly retweet conservative elites. This highlights the ideological clustering of the scientific elite in the United States.


Overall, these findings highlight the existence of a partisan divide within the scientific community, as evidenced by the differential use of emotional and moral language by both PHEs and pseudo-experts towards liberal and conservative elites. Such polarization within the public health elites has implications for the perceived credibility of health messaging, and ultimately, the ability to foster consensus and cooperation in addressing public health challenges.

\begin{figure}[!htb]
    \centering
    \includegraphics[width=\textwidth]{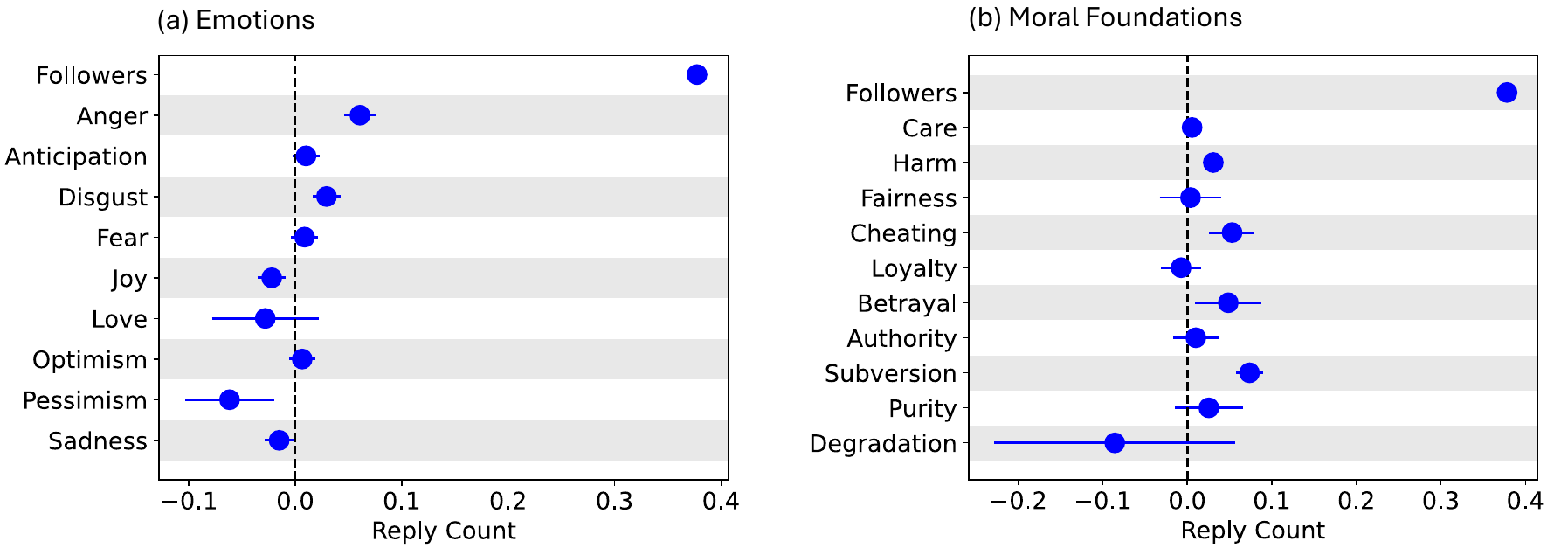}
    \caption{Engagement with PHEs. Dot and whisker plots show the regression coefficients and standard errors. The coefficients represent the increase in number of replies when an (a) emotion or (b)moral foundation is used by the PHE in the original tweet while keeping others constant.}
    \label{fig:reply_compare}
\end{figure}

\subsection*{Public Engagement with PHEs}

Effective public health communication relies not only on the dissemination of accurate information but also on how that information is received and interpreted by the public. To investigate this, we extracted original tweets (excluding retweets, reply tweets and quoted tweets) from PHEs, yielding us a corpus of $144K$ tweets. We then collect replies to these tweets, aiming to understand the engagement in response to these tweets from PHEs. However, our efforts were hindered by challenges encountered when Twitter (now X) restricted access to their academic API, limiting our ability to gather reply interactions to all original tweets. 

We were able to collect replies for $19.5K$ original PHE tweets, a total of $786K$ replies from $345K$ unique users. On average, each tweet received approximately $40.24$ replies. The distribution of replies ranged widely, from a minimum of $1$ reply to a maximum of $11.7K$ replies, with a median per tweet reply count of $5$. To quantify the effects of emotions and moral language use on the number of replies a tweet from PHEs receives, we use linear regression with the number of replies as the dependent variable and emotions or moral foundations expressed in the original tweet as the independent variables. With different PHEs having varying number of followers, the engagement their tweets garner will also be varied. In order to control for this we add the number of followers a tweet's author has as an independent variable. We execute two models, one for assessing the impact of emotions and the other for moral attitudes. The emotions model is formulated as follows:

$$
replies \sim followers + anger + anticipation + disgust + fear + joy + love + optimism + pessimism + sadness
$$


where, \verb|replies| represents the dependent variable, the number of replies to an original tweet from a PHE, \verb|followers| is the number of that PHE's followers, and \texttt{anger}, \texttt{anticipation}, \texttt{disgust}, \textit{etc.}, are binary variables indicating whether that emotion is present in the original tweet. A similar model can be written for moral foundations. 

Figure \ref{fig:reply_compare}(a) and (b) compares the impact of various emotions and moral foundations on engagement respectively. We find that, controlling for the number of followers, presence of anger and disgust in the original tweet generates more replies. This is also true for negative moral language: presence of harm, cheating, betrayal and subversion over its corresponding moral virtues is associated with more replies.  
These findings add important nuance to previous research  showing that tweets expressing negative emotions~\cite{schone2021negativity} and moral outrage~\cite{brady2017emotion} are more likely to be retweeted. Specifically, while negative emotions and moral language also receive more engagement in the form of replies, not all negative language leads to higher engagement: pessimism and sadness in the original tweets is associated with fewer replies. 
Importantly, positive language can even suppress engagement, as is the case for original PHE tweets expressing joy and love.
Supplementary Tables \ref{tab:linear_model_of_emotion} and \ref{tab:linear_model_of_morality} tabulate the results of this regression analysis. 


\begin{figure}[!htb]
    \centering
    \includegraphics[width=\textwidth]{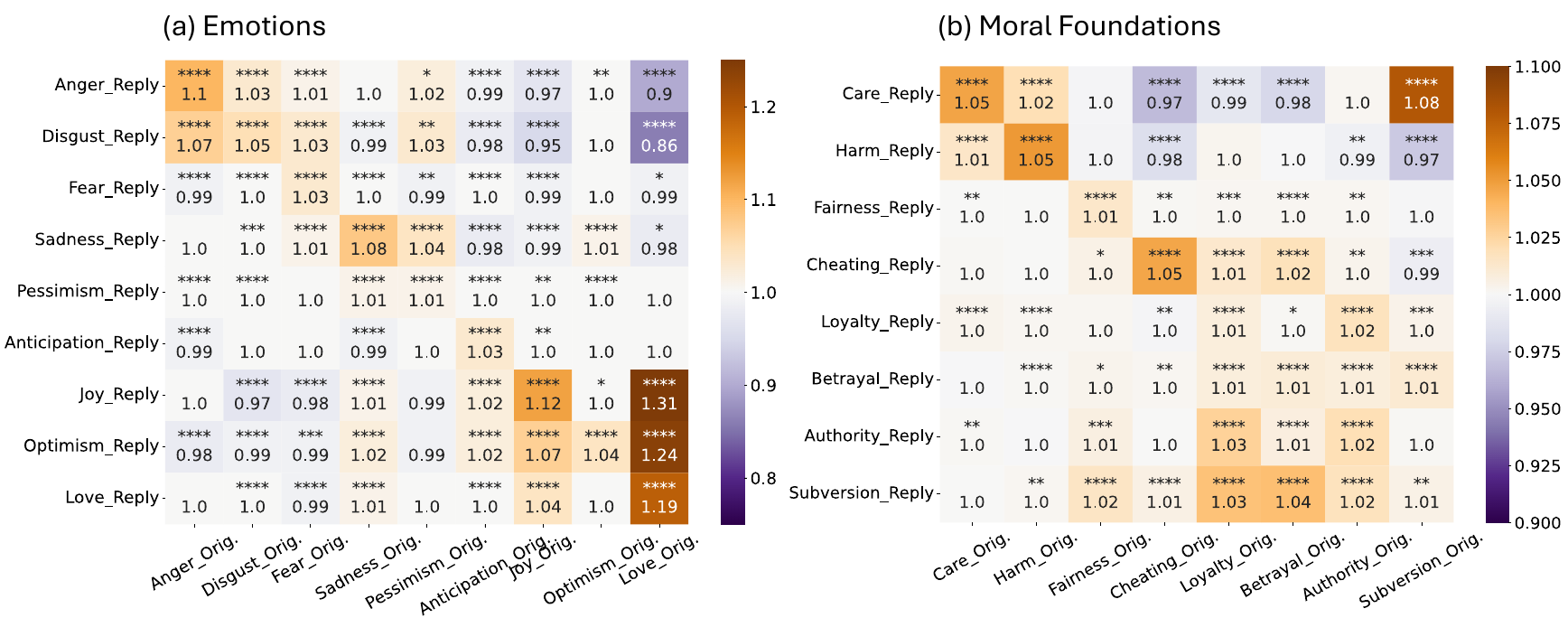}
    \caption{User engagement with PHEs and Pseudo Experts. User reactions to (a) emotional and (b) moral appeals from PHEs. Demonstrates the odds ratio of users expressing particular emotions and moral principles in response to those conveyed in the original tweets by PHEs. An odds ratio greater than 1 signifies that the user is more inclined to express an emotion/moral foundation, equal to 1 indicates equivalent odds, and less than 1 suggests a decreased likelihood of expression. * indicates significance at $p<0.05$, ** - $p<0.01$, *** - $p<0.001$, **** - $p<0.0001$.}
    \label{fig:reply_main_emotion_correspond}
\end{figure}

This leads us to question how users react to emotional and moral language of public health experts. We look at whether the use of a particular emotion (or moral foundation) in a PHE tweet triggers similar language in the replies. We use multivariate logistic regression models to quantify the odds-ratio of a user expressing an emotion or moral attitude when presented with a tweet from the PHE containing certain emotional language (similarly extended to morals):

$$
\verb|<Emotion>|_{reply} \sim anger_{orig} + disgust_{orig} + fear_{orig} + sadness_{orig} + pessimism_{orig} +\\
$$
$$
anticipation_{orig} + joy_{orig} + optimism_{orig} + love_{orig} + followers \\
$$


Figure \ref{fig:reply_main_emotion_correspond}(a) shows the odds ratio of users expressing a specific emotion in response to the emotion conveyed in a PHE's original tweet. We observe that users are more likely to express the same emotions as the original tweets. Interestingly, when PHEs express joy and love, users are more likely to express  joy, love and optimism, and less likely to express anger and disgust. Conversely, when PHEs convey negative emotions, users are more likely to express anger and disgust. Respondents generally match the emotion tone of the original tweets, except when PHEs express sadness, which respondents counter with optimism. Figure \ref{fig:reply_main_emotion_correspond}(b) illustrates the odds ratio of users expressing a specific moral foundation in response to the moral foundation conveyed in a PHE's original tweet. Similar to emotions, we observe a mirroring effect in the usage of most moral foundations between PHEs and ordinary users. Surprisingly, we notice a higher odds ratio of care being expressed when subversion is used by PHEs. 

These findings underscore the impact of emotional resonance in shaping user responses to public health messaging. When PHEs rely on positive framing, users tend to echo these positive sentiments. This suggests that the emotional and moral framing used by public health experts influences the emotion expressed by other users. Conversely, when negative framing is employed, users tend to reflect these sentiments with increased negative expressions. These insights illuminate the importance of carefully crafting public health messaging so as to foster cohesive online discourse.

\section*{Discussion}
The COVID-19 pandemic not only brought about an unprecedented global health crisis but also highlighted the critical role of effective communication in navigating public health challenges. Social media platforms, particularly Twitter, emerged as vital channels for health experts to disseminate timely and reliable information to the public. However, as the pandemic unfolded, discussions surrounding it became increasingly polarized, leading to the proliferation of misinformation and conspiracy theories, often propagated by pseudo-experts. By analyzing a substantial dataset of over $515K$ tweets generated by roughly $845$ elites, this study delves into this infodemic, comparing the emotional and moral appeals employed by public health experts and pseudo-experts on Twitter across various pandemic related issues. 

\subsection*{Principal Results}

Firstly, we categorized tweets from PHEs and pseudo-experts to seven different issues---origins of the pandemic, stay-at-home lockdown mandates, masking mandates, healthcare infrastructure, reopening the education system, therapeutics and vaccinations. Our analysis of Twitter discourse from public health experts and pseudo-experts shows that they focused on different subsets of issues, similar to what was found in other work~\cite{nogara2022disinformation}. While PHEs predominantly focused on promoting public health measures such as social distancing, masking, improving healthcare infrastructure, safer reopening of schools, pseudo-experts 
opposed lockdowns and mask mandates and promoted alternative views on therapeutics and virus origins. 

Our study goes beyond previous works to reveal emotional and moral divides surrounding key COVID-19 issues. 
While, PHEs expressed more positive emotions, and emphasize moral virtues when discussing lockdowns, masking, healthcare, and vaccines,  pseudo-experts expressed more anger and disgust in their posts on these issues and instead were more positive about therapeutics and alternative cures.
The disparate use of emotional and moral language towards ideological elites, showed that PHEs were aligned with liberal elites and pseudo-experts with conservative elites, potentially signalling the role of health influencers in growing polarization.  

Finally, our study identifies a clear trend in how people react to tweets from public health experts. When these experts express anger, disgust, or moral values, they tend to get more replies from users. Additionally, we found that it's more likely for replies to echo the same emotional or moral sentiments as in the original tweets from PHEs. 

\subsection*{Limitations and Future Work}

We note several areas of future work. Given that Twitter users are not a representative sample of the United States population, our findings may primarily reflect the perspectives of a specific demographic (i.e., younger, more liberal, better educated, more interested in politics) \cite{wojcik2019sizing}. Future studies can instead focus on multiple platforms and incorporate multimodal data. While our study examines COVID-19-related discourse, there's potential for investigations into scientific divisions in perspectives on polarized topics such as climate change and genetically modified foods. Moreover, exploring the growing debate on the factors contributing to the decline in adolescent mental health presents another avenue for inquiry.

Despite being state-of-the art models to identify emotions and moral language \cite{alhuzali2021spanemo,guo2023data}, the these models aren't oracles. The advent of more powerful, albeit expensive, instruction-tuned language models like ChatGPT, future works can leverage them at scale to identify emotions and moral attitudes more accurately. However, given these tasks' inherent ambiguity, this application might not be straightforward. We emphasize that the event related shifts in use of emotions and moral language allows us to make observational assertions rather than causal ones. Future work can attempt to run natural experiments to quantify the impact of events on different cohorts.Aside from this, the disruption in our access to Twitter's Education Access API resulted in us only being able to collect replies for a subset of the PHEs tweets in our dataset. However, it is important to note that the subset of tweets we collected replies for were not intentionally sampled or biased in any way. Finally, our dataset, despite being extensive, is between January 2020 and January 2021, limiting our findings to this period and excluding potential shifts in perspectives that occurred later on.

\subsection*{Conclusions}

In summary, our study offers valuable insights into the dynamics of public health communication on social media amidst the unfolding pandemic, exploring viewpoints from both health experts and pseudo-experts. The identification of an ideological and emotional division in the scientific community poses a potential barrier to consensus-building and undermines public trust in health messaging. Nevertheless, policymakers can leverage findings from interactions with public health experts to devise tailored strategies aimed at enhancing consensus. Tackling these obstacles demands a multifaceted approach, integrating fact-checking, debunking initiatives, and targeted communication efforts designed to cultivate trust and encourage critical thinking among the public.

\section*{Abbreviations}
PHE: Public Health Experts; 
MFT: Moral Foundations Theory;
HCQ: Hydroxychloroquine;
CDC: Center for Disease Control;
EUA: Emergency Use Authorization;
FDA: Food and Drugs Administration;
mRNA: Messenger Ribonucleic Acid

\bibliography{references}

\section*{Multimedia Appendix}

\begin{longtable}[!hb]{p{0.75in}p{2.1in}p{2in}} 
\caption{Issue Perspectives. Summary of perspectives of PHEs and pseudo-experts across issues.} \\
\toprule
Issue & PHEs & Pseudo Experts \\
\midrule
\endfirsthead

\multicolumn{3}{c}%
{{\tablename\ \thetable{} -- continued from previous page}} \\
\toprule
Issue & PHEs & Pseudo Experts \\
\midrule
\endhead

\midrule
\multicolumn{3}{r}{{Continued on next page}} \\
\endfoot

\bottomrule
\endlastfoot

Origins & Some express skepticism or criticism towards certain theories, such as the notion of an engineered origin or the lack of transparency from China regarding data sharing. Others highlight official statements or investigations, such as the World Health Organization's conclusion that it's unlikely the virus originated from a laboratory accident and the deployment of experts to further investigate the suspected outbreak zone. Additionally, there are calls for accountability and transparency in understanding the origins of the virus, as well as efforts to combat misinformation surrounding the topic.  & Some individuals and sources express skepticism about the official narrative, questioning the likelihood of the virus originating naturally and suggesting the possibility of a lab leak or intentional release. Criticism is directed towards controversial experiments in China. There are also concerns raised about potential conflicts of interest and the credibility of investigations led by organizations like the WHO. Some voices call for sanctions on Chinese research institutions.\\
\hline
Lockdowns & Tweets primarily focus on various aspects of pandemic response, including the reopening of schools, concerns about COVID-19 cases and variants, calls for continued vigilance, updates on cases, and recommendations for safety measures like mask-wearing and social distancing. & Some tweets criticize lockdowns, calling them ineffective and highlighting instances where they have been ruled unconstitutional or have failed to prevent the spread of the virus. Others question the necessity of lockdowns and suggest that they have negative effects on children's health and well-being.\\
\hline
Masking &  The tweets report on individuals testing positive for the virus and emphasize the importance of continued mask-wearing and social distancing measures to mitigate the spread of COVID-19. They mention the effectiveness of masks in preventing transmission and advocate for their widespread use, especially in areas with high transmission rates or among unvaccinated individuals. Additionally, there are mentions of policy changes regarding mask mandates in schools and public spaces, as well as discussions about the potential impact of mask-wearing on reducing COVID-19 deaths.  & The tweets present a variety of opinions and arguments regarding the effectiveness, necessity, and societal impact of wearing masks. Most tweets express skepticism about the efficacy of masks in preventing the spread of COVID-19, citing studies and personal beliefs. Others criticize mask mandates and question the motives behind enforcing them, while some highlight instances of mask enforcement or incidents related to mask-wearing. Additionally, there are mentions of government policies and public health recommendations regarding mask-wearing, as well as discussions about individual freedoms and government overreach. \\
\hline
Education & They cover various aspects such as the challenges faced by schools in reopening safely, debates surrounding COVID-19 safety protocols in schools, and the impact of the pandemic on students' education and well-being. Concerns about COVID-19 outbreaks in educational institutions, the efficacy of symptom-based screening in containing outbreaks, and the effectiveness of COVID-19 vaccines in preventing severe illness among students are also highlighted. Additionally, there are discussions about the need for innovative solutions to address the educational challenges posed by the pandemic and the importance of prioritizing students' safety and well-being in decision-making processes. & They discuss the impacts of school closures and remote learning on students, highlighting concerns about academic setbacks, mental health issues, and the effectiveness of virtual education. Debates arise regarding the reopening of schools, with differing opinions on the risks involved and the appropriate measures needed to ensure safety. Additionally, there are critiques of teachers' unions, government policies, and media coverage related to education during the pandemic. Overall, the tweets reflect the complex and polarizing discussions surrounding education and online schooling amidst the ongoing public health crisis. \\
\hline
Healthcare & Some emphasize the effectiveness of vaccines in reducing the risk of severe illness and death, while others discuss concerns about long COVID and the potential impact on public health. There are also mentions of challenges faced by healthcare workers, such as infections among medical staff and the importance of vaccinating frontline providers. Additionally, the tweets touch on issues related to pandemic communication, hospital capacity, and vaccination efforts in different regions. & Some tweets highlight success stories and alternative treatments, such as Mexico City's distribution of IVM kits and Swiss doctor Klaus Schustereder's home treatment approach. Others express skepticism or criticism of mainstream approaches, including concerns about vaccine efficacy, medical ethics violations, and government directives. There are also mentions of specific incidents, such as lawsuits over clinical trial reactions and the controversial directive in New York to return COVID-positive patients to nursing homes. \\
\hline
Therapeutics & These tweets caution against their widespread adoption due to insufficient scientific support and potential risks. Instances of misinformation and controversy, such as the promotion of unproven treatments like oleandrin and the widespread use of ivermectin in veterinary medicine, are also highlighted. Additionally, there are warnings against self-administering medications like dexamethasone without medical supervision, as well as calls for responsible reporting and adherence to evidence-based practices in treating COVID-19. &  Some sources advocate for their use, citing studies and trials that suggest their effectiveness in treating COVID-19. They argue that HCQ, when combined with zinc and azithromycin, has shown positive outcomes in reducing hospitalizations and mortality rates. Similarly, ivermectin is promoted as a potential treatment, with proponents highlighting its benefits in preventing and treating COVID-19 infections, particularly when administered alongside other medications. These proponents criticize media censorship and bureaucratic obstruction of these treatments, emphasizing the need for further research and widespread adoption. \\
\hline
Vaccines &The tweets cover a wide range of perspectives on COVID-19 vaccines, including discussions about vaccine effectiveness, breakthrough cases, concerns about vaccine distribution, and misinformation. They highlight instances of vaccine hesitancy, breakthrough infections among unvaccinated individuals, and the importance of vaccination for pregnant individuals. Additionally, there are discussions about vaccine mandates for healthcare workers and the potential need for booster vaccines. The tweets also address concerns about vaccine safety, including debunking myths about vaccine impact on fertility and clarifying the absence of evidence regarding vaccine-related adverse effects. & The tweets cover a wide range of perspectives and concerns about vaccines, including adverse reactions, skepticism about their effectiveness and safety, potential long-term risks, concerns about coercion, and issues with data integrity in clinical trials. Some highlight specific incidents of adverse reactions or alleged vaccine-related deaths, while others express skepticism about the motives of pharmaceutical companies and government policies regarding vaccination. There's also discussion about the efficacy of vaccines, potential side effects, and ongoing research and testing. \\
\label{tab:perspec}
\end{longtable}

 To provide a concise summary of the diverse perspectives underlying emotions expressed by health and pseudo-experts in response to significant events, we randomly select 25 tweets for each emotion within a 15-day window before and after the event. The objective is twofold: i) compare the share of tweets expressing a particular emotion 15 days before and after the event, and ii) explain the perspectives contributing to these fluctuations. We prompt ChatGPT as follows:

\begin{quote}
\verb|What <Emotion> is being expressed in these tweets:<T>|
\end{quote}

where, \verb|<Emotion>| can take values from \verb|[Anticipation, Joy, Optimism, Anger, Disgust, Sadness, Fear]|, and \verb|<T>| denotes a concatenation of the 25 randomly sampled tweets for each \verb|<Emotion>| and group pair. The results, presented in Table \ref{tab:emotions_summ}, depict the change in the share of tweets expressing an emotion for the PHEs ($\Delta$ Health) and pseudo-experts ($\Delta$ Pseudo) around each event. It is important to note that this method explains what's driving an increase in tweeting activity but cannot explain declines in activity.


\begin{longtable}[!ht]{p{0.4in}p{0.4in}p{0.3in}p{0.3in}p{1.75in}p{1.75in}}
\caption{Summary of emotions. Summary of emotions expressed by PHEs and pseudo-experts in response to major events.}\\
\toprule
Date & Emotion & $\Delta$ Health &$\Delta$ Pseudo & Summ. PHE & Summ. Pseudo \\
\midrule
\endfirsthead

\multicolumn{6}{c}%
{{\bfseries \tablename\ \thetable{} -- continued from previous page}} \\
\toprule
Date & Emotion & $\Delta$ Health &$\Delta$ Pseudo & Summ. PHE & Summ. Pseudo \\
\midrule
\endhead

\midrule
\multicolumn{6}{r}{{Continued on next page}} \\
\midrule
\endfoot

\bottomrule
\endlastfoot

2020-03-15 &  joy & \textcolor{ForestGreen}{0.06} &\textcolor{ForestGreen}{0.04} &     The joy expressed in this text is primarily about the collective efforts and progress made in response to the COVID-19 pandemic. It includes updates on various initiatives such as online learning courses for students amidst school closures, advancements in COVID-19 testing processes at hospitals, appreciation for scientists and physicians contributing to public understanding, transparent reporting of testing data by UWVirology, and community outreach efforts to underserved populations. Overall, it showcases a sense of resilience, collaboration, and dedication in combating the pandemic. &      The joy expressed here primarily revolves around positive developments in combating the COVID-19 pandemic. This includes excitement about President Trump's criteria for assessing the threat level by county, France's approval of chloroquine after successful patient recovery, and a large study indicating chloroquine's effectiveness against COVID-19. Additionally, there's gratitude towards healthcare workers, admiration for community resilience, and a sense of humor amidst the challenging circumstances. \\
2020-03-15 &      optimism &      \textcolor{ForestGreen}{0.09} &\textcolor{ForestGreen}{0.02} &       The optimism expressed here revolves around the belief that collective action, such as staying at home and implementing mitigation strategies, can effectively control the spread of COVID-19. There's praise for successful responses in places like South Korea, as well as calls for clear leadership and guidance from figures like Governor Cuomo. Additionally, there's hope for the future, with discussions about potential referendums and confidence in Joe Biden's leadership in addressing the pandemic. &    ---- \\
2020-03-15 &       disgust &      \textcolor{ForestGreen}{0.04} &\textcolor{ForestGreen}{0.10} & The disgust expressed here is directed towards several aspects of the COVID-19 pandemic response. There's frustration over the prioritization of celebrities for testing while hospitalized patients struggle to access tests. Additionally, there's anger towards governmental actions and failures, such as the undercutting of oversight provisions in relief bills and the perceived lack of commitment to essential steps in controlling the pandemic. &    The disgust expressed here is directed towards various entities and actions related to the COVID-19 pandemic response. There's frustration towards the mishandling of the crisis by public officials, including the Massachusetts Governor's delay in issuing a shelter-in-place order. There's also criticism of Senate Democrats for allegedly leveraging relief bills. Additionally, there's concern about misinformation regarding the transmission of the virus through cash transactions, as well as skepticism towards the World Health Organization's communication on the matter. The mention of "CCPVirus" suggests broader disdain towards China and its handling of the outbreak. \\
2020-03-15 &anger &      \textcolor{ForestGreen}{0.03} &\textcolor{ForestGreen}{0.11} &        The anger is primarily directed at various governmental entities and officials. Specifically, frustration is expressed towards Governor Kemp of Georgia for not implementing stricter measures to combat COVID-19, despite expert advice. Additionally, Florida Sheriff Chad Chronister's decision to seek an arrest warrant for a pastor who disregarded coronavirus orders, described as one of Trump's friends who visited the White House, is criticized. There's also discontent regarding Senator Burr's stock trades and the lack of transparency from the CDC regarding their coronavirus communications plan. &        Firstly, there's frustration with New York Governor Cuomo over the handling of public transportation during the COVID-19 pandemic, which is seen as contributing to the spread of the virus. Criticism is also aimed at the World Health Organization (WHO) for its perceived bias towards China, with calls for a shakeup of its leadership. Additionally, there's skepticism regarding the motives behind extending the lockdown till the end of April, with concerns about its economic and societal impact. \\
\hline
2020-11-09 &  anticipate &      \textcolor{ForestGreen}{0.03} &\textcolor{ForestGreen}{0.03} &       The anticipation here is centered around the development and potential approval of COVID-19 vaccines. Specific entities mentioned include Pfizer and Moderna, with positive news regarding vaccine efficacy generating optimism. There's also anticipation for discussions on COVID-19 vaccine development and pandemic preparedness at events like the G20 summit, highlighting the global interest and efforts in combating the pandemic.  &     There's a sense of urgency to expedite Operation Warp Speed, aimed at accelerating the development and distribution of COVID-19 vaccines. Furthermore, the imminent arrival and distribution of Lilly's monoclonal antibody drug for COVID patients raise hopes for its potential to reduce hospitalizations and expedite recovery. Lastly, events aimed at raising awareness about vaccine injury and discussions regarding COVID-19 vaccines contribute to the overall anticipation surrounding vaccine development and distribution. \\
2020-11-09 &  joy &      \textcolor{ForestGreen}{0.03} &\textcolor{Gray}{0.00} &The joy expressed here is centered around the positive developments regarding COVID-19, particularly the news from Pfizer about the effectiveness of its coronavirus vaccine. With early data suggesting more than 90\% efficacy and fewer infections among participants who received the vaccine compared to those who received a placebo, there's a sense of excitement and optimism. Additionally, the FDA's emergency approval of the first rapid coronavirus test for home use in just 30 minutes is seen as a significant step forward in combating the pandemic. &    ---- \\
2020-11-09 &      optimism &      \textcolor{ForestGreen}{0.04} &\textcolor{ForestGreen}{0.03} &  Firstly, there's anticipation surrounding the potential arrival of a vaccine within the next few months, which is seen as a remarkable feat in combating the pandemic. Additionally, there's hope stemming from a new study suggesting that immunity to the coronavirus might last for years, providing a more optimistic outlook on long-term protection against the virus. Furthermore, the formation of a COVID-19 task force by President-elect Biden, the promising efficacy of vaccine candidates, and the FDA's emergency use authorization for monoclonal antibodies are viewed as significant steps forward in addressing the crisis. &        The success of Operation Warp Speed, particularly the announcement of Moderna's COVID-19 vaccine candidate being nearly 95\% effective, is celebrated as a significant milestone in saving lives and overcoming the pandemic. Furthermore, there's appreciation for healthcare professionals and their efforts, as evidenced by Dr. Jeff Barke's advocacy for safely reopening the country and the FDA's emergency use authorization for Lilly's monoclonal antibody drug.  \\
2020-11-09 &       disgust &     \textcolor{Red}{-0.06} &        \textcolor{Red}{-0.07} &---- &    ---- \\
2020-11-09 &anger &     \textcolor{Red}{-0.07} &        \textcolor{Red}{-0.06} &---- &    ---- \\
\bottomrule
\label{tab:emotions_summ}
\end{longtable}

To provide a concise summary of the diverse perspectives underlying moral attitudes expressed in response to significant events, we randomly select 25 tweets for each moral foundation in a 15-day window before and after the event. The objective as with emotions is twofold: i) to compare the proportion of tweets expressing a particular moral foundation 15 days before and after the event, and ii) to comprehend the perspectives contributing to these fluctuations. To evaluate the latter, we prompt ChatGPT as follows:

\begin{quote}
\verb|What <Moral Foundation> is being expressed in these tweets:<T>|
\end{quote}

where, \verb|<Moral Foundation>| can take values from \verb|[Care, Harm, Fairness, Cheating, Authority,| \\
\verb|Subversion, Loyalty, Betrayal, Purity, Degradation]|, and \verb|<T>| denotes a concatenation of the 25 randomly sampled tweets for each \verb|<Moral Foundation>| and group pair. The results presented in Table \ref{tab:mft_summ} depict the change in the proportion of tweets expressing a moral foundation for PHEs ($\Delta$ PHE) and pseudo-experts ($\Delta$ Pseudo) around each event. 

\begin{longtable}[!ht] {p{0.4in}p{0.4in}p{0.3in}p{0.3in}p{1.75in}p{1.75in}}
\caption{Summary of moral attitudes expressed by PHEs and pseudo-experts in response to major events.}\\
\toprule
Date & Emotion & $\Delta$ Health &$\Delta$ Pseudo & Summ. PHE & Summ. Pseudo \\
\midrule
\endfirsthead
\multicolumn{6}{c}%
{{\bfseries \tablename\ \thetable{} -- continued from previous page}} \\
\toprule
Date & MFT & $\Delta$ Health &$\Delta$ Pseudo & Summ. PHE & Summ. Pseudo \\
\midrule
\endhead

\midrule
\multicolumn{6}{r}{{Continued on next page}} \\
\midrule
\endfoot

\bottomrule
\endlastfoot

2020-03-15 &   care &\textcolor{ForestGreen}{0.05} &    \textcolor{Red}{-0.01} &    They convey concern about governmental decisions regarding ventilator distribution, gratitude towards healthcare workers for their dedicated care, and acknowledgment of the effectiveness of measures such as staying at home. Additionally, there's advocacy for widespread testing, educational resources for children, and critiques of media response and relief proposals that exclude low-income individuals. Overall, these messages emphasize the importance of prioritizing public health, supporting frontline workers, and advocating for effective leadership and inclusive relief efforts during the ongoing crisis. &  ---- \\
2020-03-15 &   harm &    \textcolor{Red}{-0.04} &    \textcolor{Red}{-0.02} & ---- &  ---- \\
\hline
2020-11-09 &   care &\textcolor{ForestGreen}{0.05} &\textcolor{ForestGreen}{0.06} &  There's care expressed in advocacy for safety precautions in schools and outdoor activities to ensure the well-being of students and staff. Following, vaccine trials meeting efficacy endpoints, there is optimism and care towards finding effective solutions to combat the virus and protect individuals globally. &  In addition to vaccines, attention is given to scientific studies linking the MMR vaccine to potential protection against COVID-19, demonstrating a commitment to understanding potential preventive measures. The announcement of Pfizer's COVID-19 vaccine's efficacy in clinical trials reflects optimism and care towards finding effective vaccines to combat the virus. However, there are concerns expressed regarding vaccines, particularly due to the requirement of ultra-cold storage for Pfizer's vaccine and the novel mRNA technology indicating a need for careful planning and adoption. Bill Gates' plan for a potentially mandatory vaccine, along with the suggestion that individuals may need a "digital certificate" of vaccination to resume normal life, is met with criticism and perceived as unpopular \\
2020-11-09 &   harm &    \textcolor{Red}{-0.04} &    \textcolor{Red}{-0.03} & ---- &  ---- \\
\bottomrule
\label{tab:mft_summ}
\end{longtable}

\begin{table*}[!ht]
    \centering
    \begin{tabular}{l|c|c|c}
         \textbf{Variable}&  \textbf{Coefficient} & \textbf{Std.Err} & \textbf{$P>|t|$}\\\hline 
         constant & -2.8764 & 0.069 & 0.000**\\
         log(\# followers) & 0.4345 & 0.006 & 0.000**\\
         anger & 0.1636 & 0.039 & 0.000**\\
         anticipation & 0.0344 & 0.033& 0.296\\
         disgust & 0.0730 &0.034 & 0.033*\\
         fear & 0.0239 & 0.033 & 0.473\\
         joy & -0.0472 & 0.035& 0.180\\
         love & -0.0914 & 0.133& 0.491\\
         optimism &  0.0257 & 0.032 & 0.429\\
         pessimism & -0.1447 & 0.111& 0.193\\
         sadness & -0.0383 & 0.035 & 0.278\\
         surprise & 0.5471 & 0.134 & 0.000**\\
         trust & -0.3276 & 0.158 & 0.038*\\
    \end{tabular}
    \caption{Emotions and Engagement. Linear model predicting log(\# replies). The Adj. R-squared of the model is 0.204.}
    \label{tab:linear_model_of_emotion}
\end{table*}

\begin{table*}[ht]
    \centering
    \begin{tabular}{l|c|c|c}
         \textbf{Variable}&  \textbf{Coefficient} & \textbf{Std.Err} & \textbf{$P>|t|$}\\\hline 
         constant & -2.8736 & 0.069& 0.000**\\
         log(\# followers) & 0.4353 & 0.006& 0.000**\\
         care & 0.0093 &  0.028& 0.737\\
         harm & 0.0769 & 0.026 & 0.003**\\
         fairness & 0.0021 & 0.096 & 0.982\\
         cheating & 0.1267 & 0.071 & 0.075\\
         loyalty & -0.0162 & 0.063& 0.795\\
         betrayal & 0.1146 & 0.104& 0.270\\
         authority & 0.0242 & 0.071& 0.734\\
         subversion & 0.1886 & 0.042 & 0.000**\\
         purity & 0.0804 & 0.106 & 0.448\\
         degradation & -0.1935 & 0.377 & 0.608\\
    \end{tabular}
    \caption{Moral Foundations and Engagement. Linear model predicting log(\# replies). The Adj. R-squared of the model is 0.201.}
    \label{tab:linear_model_of_morality}
\end{table*}

\end{document}